\documentclass[10pt,twocolumn,letterpaper]{article}

\usepackage[]{iccv}      
\usepackage[accsupp]{axessibility}  

\definecolor{iccvblue}{rgb}{0.21,0.49,0.74}
\usepackage[pagebackref,breaklinks,colorlinks,allcolors=iccvblue]{hyperref}

\usepackage{makecell}
\usepackage{graphicx}
\usepackage{hyperref}
\usepackage{multirow} 
\usepackage{float} 
\usepackage{amsmath} 
\usepackage{adjustbox} 
\usepackage{verbatim} 
\usepackage{hwemoji}
\usepackage{pifont} 
\usepackage[normalem]{ulem}
\useunder{\uline}{\ul}{}
\usepackage{color, colortbl,booktabs}
\definecolor{Gray}{gray}{0.9}
\newcolumntype{a}{>{\columncolor{Gray}}c}
\usepackage{array}
\usepackage{textcomp}

\ExplSyntaxOn
\NewDocumentCommand{\xhrulefill}{O{}}
 {
  \group_begin:
  \severin_xhrulefill:n { #1 }
  \group_end:
 }

\keys_define:nn { severin/xhrulefill }
 {
  height .dim_set:N    = \l_severin_xhrule_height_dim,
  thickness .dim_set:N = \l_severin_xhrule_thickness_dim,
  fill .skip_set:N     = \l_severin_xhrule_fill_skip,
  height .initial:n    = 0pt,
  thickness .initial:n = 0.4pt,
  fill .initial:n      = 0pt plus 1fill,
 }

\cs_new_protected:Nn \severin_xhrulefill:n
 {
  \keys_set:nn { severin/xhrulefill } { #1 }
  \leavevmode
  \leaders\hrule 
    height \dim_eval:n { \l_severin_xhrule_thickness_dim + \l_severin_xhrule_height_dim }
    depth  \dim_eval:n { -\l_severin_xhrule_height_dim }
  \skip_horizontal:N \l_severin_xhrule_fill_skip
  \kern 0pt
}
\ExplSyntaxOff

\def\paperID{*****} 
\def\confName{ICCV}
\def\confYear{2025}

\title{\textsc{VolDoGer}: LLM-assisted Datasets for Domain Generalization in Vision-Language Tasks}

\author{Juhwan Choi\\
AITRICS\\
Seoul, Republic of Korea\\
{\tt\small jhchoi@aitrics.com}
\and
Junehyoung Kwon, Jungmin Yun, Seunguk Yu, YoungBin Kim\\
Chung-Ang University\\
Seoul, Republic of Korea\\
{\tt\small ybkim85@cau.ac.kr}
}

\begin{document}
\maketitle
\begin{abstract}
Domain generalizability is a crucial aspect of a deep learning model since it determines the capability of the model to perform well on data from unseen domains. However, research on the domain generalizability of deep learning models for vision-language tasks remains limited, primarily because of the lack of required datasets. To address these challenges, we propose \textsc{VolDoGer}: \textbf{V}isi\textbf{o}n-\textbf{L}anguage Dataset for \textbf{Do}main \textbf{Ge}ne\textbf{r}alization, a dedicated dataset designed for domain generalization that addresses three vision-language tasks: image captioning, visual question answering, and visual entailment. We constructed \textsc{VolDoGer} by extending LLM-based data annotation techniques to vision-language tasks, thereby alleviating the burden of recruiting human annotators. We evaluated the domain generalizability of various models through \textsc{VolDoGer}.
\end{abstract}    
\section{Introduction}
\label{sec:intro}

Vision-language models have evolved and demonstrated outstanding performance in various tasks \cite{chen2023vlp} such as image captioning \cite{stefanini2022show}, visual question answering (VQA) \cite{wu2017visual, de2023visual}, and visual entailment (VE) \cite{xie2019visual}. However, these vision-language models can suffer from domain shift, which is a significant challenge for deep learning models \cite{wang2018deep, fang2024source}. Domain shift refers to a phenomenon in which the domain of the data changes between the training and inference phases of a model. For example, an image classification model trained on photos may not perform well when applied to sketch images \cite{zhou2022domain}. This issue is prevalent in NLP tasks \cite{elsahar2019annotate, ramponi2020neural, calderon2023measuring} to vision-language tasks \cite{chen2017show, zhao2017dual, yang2018multitask, zhao2020cross}.

\begin{figure}[t!]
    \centering
    \includegraphics[width=\columnwidth]{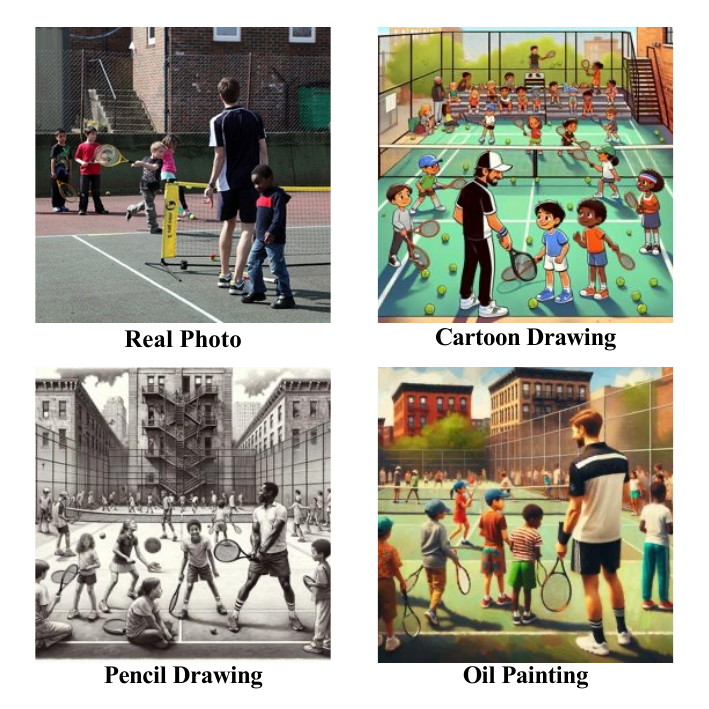}
    \caption{Examples of images with various styles in \textsc{VolDoGer}. Please refer to Appendix~\ref{app:additional-examples-annotation} for more examples with annotation.}
\label{fig:intro}
\end{figure}

Extensive research has been conducted on domain generalization to mitigate domain shift \cite{zhou2022domain, wang2022generalizing}. These lines of study aim to utilize multiple source domains to enhance the generalizability of the model against out-of-domain target domains. However, the difficulty of collecting annotated data from various source domains may diminish the practicality of domain generalization. Although it is relatively simple to gather data for unimodal tasks such as image classification or text classification \cite{blitzer2007biographies, peng2019moment}, it may be more difficult to collect data for multimodal tasks because they require a \textit{pair} of data in each modality.

Consequently, there is a lack of datasets for domain generalization in multimodal tasks, including vision-language tasks. The absence of a dedicated dataset makes it difficult to explore domain generalization in vision-language tasks. For example, existing studies on domain generalization for image captioning merged multiple existing datasets with different subjects, each of which contains real photos as input images \cite{ren2023crossing}. However, this approach fails to fully consider the diversity of domains because it contains only real photographs, thereby not accounting for the domain shift in the \textit{style} of the input image. Furthermore, recent advancements in generative models have led to a significant increase in the volume of generated content encompassing a diverse array of styles. In view of this challenge, vision-language models must be capable of delivering accurate and consistent results on generated images with various styles as well, considering that generative models can easily produce images in designated styles \cite{zhang2023text}. Hence, a specialized dataset for domain generalization in vision-language tasks is required to address these challenges and ensure a robust performance across diverse image styles.

However, it is difficult to construct such a dataset through the collection and annotation by human annotators. Unlike relatively straightforward tasks such as image classification, where images with various styles can be collected with a simple search (e.g., ``aeroplane painting'') \cite{peng2019moment}, creating a vision-language task dataset for domain generalization imposes more severe restrictions. For instance, a dataset for domain generalization in image captioning tasks would require a large set of similar images in different styles, such as cartoons, paintings, and sketches, as well as their descriptions. Moreover, these tasks require more complex human annotation procedures than simple tasks, leading to higher annotation costs and more efforts for quality control \cite{rashtchian2010collecting}.

To address these challenges and effectively construct datasets for domain generalization in vision-language tasks, we propose leveraging large language model (LLM)-based data annotation \cite{tan2024large}. LLM-based data annotation uses LLMs as data annotators to replace human annotators. Researchers have found this strategy to be cost-effective in producing consistent results compared with human annotators \cite{wang2021want, ding2023gpt}. However, previous studies on LLM-based data annotation have primarily focused on text data \cite{li2023coannotating, zhang2023llmaaa, he2023annollm, bansal2023large}. Although recent studies have applied LLM-based data annotation to image captioning tasks, they have not considered image data and relied solely on text input \cite{choi2024gpts}. In this study, we leverage recent advancements in LLM with improved image interpretation capabilities, such as GPT-4o \cite{openai2023gpt, openai2024gpt4o}, and explore the use of LLMs as multimodal data annotators by collaborating with recent image generation models \cite{betker2023improving, esser2024scaling}.

Using the proposed multimodal LLM-based data annotation, we constructed \textsc{VolDoGer}: \textbf{V}isi\textbf{o}n-\textbf{L}anguage Dataset for \textbf{Do}main \textbf{Ge}ne\textbf{r}alization, which is the first dedicated dataset designed to facilitate domain generalization across three vision-language tasks: image captioning, VQA, and VE. \textsc{VolDoGer} involves four different styles, which are real photos, cartoon drawings, pencil drawings, and oil paintings. Figure~\ref{fig:intro} showcases an example of image with various styles consisting \textsc{VolDoGer}. Based on these source data encompassing various styles, it is possible to train a model with improved domain generalizability using \textsc{VolDoGer}. In this study, we utilized \textsc{VolDoGer} to validate the presence of domain shifts in these tasks and to evaluate the effectiveness of existing domain generalization techniques.

Our contributions can be summarized as follows:
\begin{itemize}
\item To the best of our knowledge, this is one of the first work to establish multimodal LLM-based data annotation while considering multimodal inputs.
\item We release \textsc{VolDoGer}, a first dedicated dataset designed to advance research on domain generalization across three vision-language tasks.
\item Our extensive experiments demonstrate the presence of domain shift and the effectiveness of domain generalization techniques in vision-language tasks.
\end{itemize}

\section{Related Work}
\label{sec:related_work}

\subsection{Domain Generalization for Vision-Language Tasks}

Despite the lack of a dedicated dataset for domain generalization in vision-language tasks, researchers are increasingly exploring this area. For example, a relevant study proposed a framework for domain generalization in image captioning \cite{ren2023crossing}. They incorporated the use of text data through visual word guidance and sentence similarity based on previous research \cite{wang2020learning}. However, although they proposed an effective framework, the datasets they used, such as MSCOCO \cite{chen2015microsoft} and Flickr30k \cite{young2014image} exhibited significant overlap. Furthermore, these datasets exhibit limited differences in visual features because they primarily consist of real photos. In contrast, our objective is to create datasets for various vision-language tasks, such as image captioning, that encompass diverse visual styles within images.

In the field of VQA, VQA-GEN \cite{unni2023vqa} suggested constructing a dataset for domain generalization by modifying an existing dataset, which aligns with the purpose of our work. However, their manipulation strategies for visual features mostly consist of simple noise-based modifications, such as injecting blurs. Moreover, VQA-GEN is not publicly available, which reduces the usability of this study for future work. Another line of research proposed a methodology that enables task generalization on VQA datasets that require image understanding and compositional reasoning \cite{shrestha2019answer, gamage2021improved}.

To the best of our knowledge, this is the first study to explore domain generalization in visual entailment. Moreover, we propose a multimodal LLM-based data annotation pipeline and introduce \textsc{VolDoGer}, which is a publicly available dataset constructed using our pipeline, to facilitate future advancements in domain generalization for vision-language tasks.

\subsection{LLM-based Data Annotation}

As LLMs exhibit various capabilities, researchers have explored leveraging them as data annotators to replace human annotators. For example, automatic annotation through GPT-3 has demonstrated superior downstream model performance compared with human performance at a lower cost \cite{wang2021want, gilardi2023chatgpt}. Furthermore, the capability of GPT-3 to generate labeled data from scratch was demonstrated \cite{ding2023gpt}. Consequently, the exploration of LLM applications as data annotators continues to expand, underscoring their utility in streamlining and optimizing the data annotation process \cite{li2023coannotating, zhang2023llmaaa, he2023annollm, bansal2023large}.

A recent study closely related to our work also suggested data annotation for image captioning tasks, one of the tasks that we address \cite{choi2024gpts}. However, they made limited use of multimodal data since they did not consider image inputs while annotating the given data. Instead, they only paraphrase the given text input and translate the paraphrases into another language. By contrast, we aim to actively utilize LLMs as multimodal data annotators.
\section{LLM-based Data Annotation for Vision-Language Tasks}
\label{sec:annotation}

\begin{figure*}[ht]
    \centerline{
    \includegraphics[width=\textwidth]{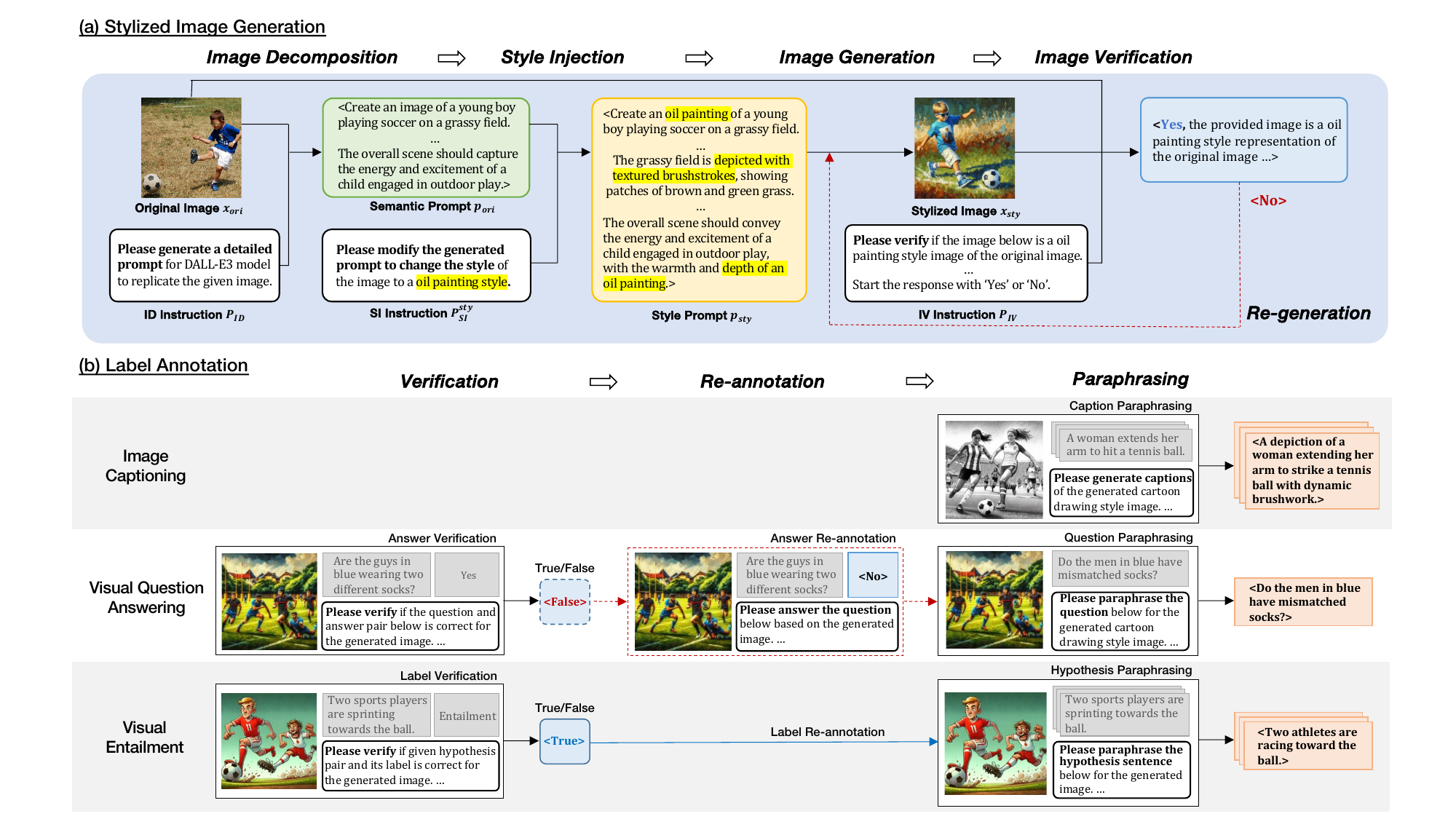}}
    \caption{Overall procedure for annotating data through our proposed framework. (a) illustrates the process for stylized image generation in Section~\ref{sec:annotation-image}, and (b) displays the label annotation process for each task, which is described in Section~\ref{sec:annotation-label-cap},~\ref{sec:annotation-label-vqa}, and~\ref{sec:annotation-label-ve}.}
\label{fig:framework}
\end{figure*}

In this section, we introduce the proposed framework for multimodal LLM-based data annotation for vision-language tasks, as illustrated in Figure~\ref{fig:framework}. The framework comprises two primary phases: stylized image generation and label annotation. Although the stylized image generation process was shared across the three tasks, the label annotation process varied slightly to accommodate the specific characteristics of each task. The objective of this framework is to convert the given dataset $\mathcal{D}_{\textit{ori}}$ into a transferred dataset with the designated style $\mathcal{D}_{\textit{sty}}$, where the input image $x_{\textit{ori}}$ is transformed into stylized image $x_{\textit{sty}}$. We utilize a multimodal LLM $\mathcal{M}$ to perform this transformation. For the exact instruction prompts, please refer to Appendix~\ref{app:prompt-data-annotation}.

\subsection{Stylized Image Generation}
\label{sec:annotation-image}

In the first phase, the framework aims to create an image $x_{\textit{sty}}$ that retains the content and semantics of $x_{\textit{ori}}$ but has a designated style. This phase consisted of four steps: image decomposition, style injection, image generation, and image verification.

\noindent \textbf{Image Decomposition.} We first input $x_{\textit{ori}}$ from $\mathcal{D_{\textit{ori}}}$ into $\mathcal{M}$ with the instruction $P_{\textit{ID}}$ to generate a prompt describing semantics in $x_{\textit{ori}}$, $p_{\textit{ori}} = \mathcal{M}(P_{\textit{ID}}, x_{\textit{ori}})$, which can be used to reconstruct $x_{\textit{ori}}$ through an image generation model $\mathcal{G}$.

\noindent \textbf{Style Injection.} Next, we transform $p_{\textit{ori}}$ into a stylized prompt $p_{\textit{sty}} = \mathcal{M}(P^{\textit{sty}}_{\textit{SI}}, p_{\textit{ori}})$ based on instruction $P_{\textit{SI}}$. The generated $p_{\textit{sty}}$ retains the content of $x_{\textit{ori}}$ while incorporating information about the desired style.

\noindent \textbf{Image Generation.} In this step, we pass the stylized prompt $p_{\textit{sty}}$ to the text-to-image generation model $\mathcal{G}$. Subsequently, a transformed image with the desired style $x_{\textit{sty}}$ is generated by $x_{\textit{sty}} = \mathcal{G}(p_{\textit{sty}})$. Appendix~\ref{app:additional-examples-annotation-prompt} provides the generated image $x_{\textit{sty}}$ and its prompt $p_{\textit{sty}}$.

\noindent \textbf{Image Verification.} It is important to note that the generated $x_{\textit{sty}}$ may not fully capture the core semantics of $x_{\textit{ori}}$. The distinction between $x_{\textit{ori}}$ and $x_{\textit{sty}}$ could complicate the subsequent annotation process; that is, the original label may not correspond to $x_{\textit{sty}}$ if it deviates significantly from $x_{\textit{ori}}$.

To address this issue, we propose introducing an image verification step for the generated $x_{\textit{sty}}$. We simultaneously pass $x_{\textit{sty}}$ and $x_{\textit{ori}}$ to $\mathcal{M}$ and compare the two images to verify whether $x_{\textit{sty}}$ retains the essential content of $x_{\textit{ori}}$. This is formulated as $v_{\textit{IV}} = \mathcal{M}(P_{\textit{IV}}, x_{\textit{ori}}, x_{\textit{sty}})$. If the candidate $x_{\textit{sty}}$ does not pass the verification; in other words, if $v_{\textit{IV}}$ is false,  we return to the image generation step and create a new $x_{\textit{sty}}$. This verification process is crucial for ensuring the accuracy and consistency of $x_{\textit{sty}}$ relative to $x_{\textit{ori}}$, thereby maintaining the validity of the annotations.

After completing the verification process, we proceed to the next phase: label annotation.

\subsection{Label Annotation for Image Captioning Task}
\label{sec:annotation-label-cap}

The image captioning task aims to generate a description $y$ from a given image $x$. Unlike the other two tasks, the image captioning task does not require any additional input besides the image. Our goal is to create a data pair $d_{\textit{sty}} = (x_{\textit{sty}}, y_{\textit{sty}})$ for the task.

\noindent \textbf{Caption Paraphrasing.} Instead of directly assigning the original $y_{\textit{ori}}$ to the generated $x_{\textit{sty}}$, we generate a paraphrase of $y_{\textit{ori}}$ as $y_{\textit{sty}}$, while considering the style of $x_{\textit{sty}}$. This process is beneficial for offering the model diverse expressions \cite{fan2023improving}. It is also crucial to avoid duplicating the label data, which can negatively impact the training procedure \cite{schofield2017quantifying}. To this end, we pass $x_{\textit{sty}}$ and $y_{\textit{ori}}$ to $\mathcal{M}$, obtaining $y_{\textit{sty}} = \mathcal{M}(P_{\textit{CP}}, x_{\textit{sty}}, y_{\textit{ori}})$.

\subsection{Label Annotation for Visual Question Answering Task}
\label{sec:annotation-label-vqa}

The purpose of VQA task is to answer question $q$ based on the given image $x$. The VQA model takes $x$ and $q$ as inputs and predicts the answer $y$. We aim to create a data pair for the VQA task as $d_{\textit{sty}} = (x_{\textit{sty}},  q_{\textit{sty}}, y_{\textit{sty}})$.

\noindent \textbf{Answer Verification.} Although $x_{\textit{sty}}$ may pass the image verification step, the original label $y_{\textit{ori}}$ for question $q_{\textit{ori}}$ may not be valid for $x_{\textit{sty}}$ owing to minor differences. For instance, if the question $q_{\textit{ori}}$ was ``How many cups are on the table?'' and $x_{\textit{ori}}$ had two cups, but $x_{\textit{sty}}$ contains four cups, the original label $y_{\textit{ori}}$ ``two'' would no longer be valid for $x_{\textit{sty}}$. 

To verify $y_{\textit{ori}}$ with respect to $x_{\textit{sty}}$, we utilize $\mathcal{M}$ to confirm if $y_{\textit{ori}}$ is the answer to $q_{\textit{ori}}$ given $x_{\textit{sty}}$ based on $v_{\textit{AV}} = \mathcal{M}(P_{\textit{AV}}, x_{\textit{sty}}, q_{\textit{ori}}, y_{\textit{ori}})$. If $y_{\textit{ori}}$ is a valid answer to $q_{\textit{ori}}$ given $x_{\textit{sty}}$, we assign $y_{\textit{ori}}$ as $y_{\textit{sty}}$. Otherwise, if $y_{\textit{ori}}$ is not a valid answer for $q_{\textit{ori}}$ given $x_{\textit{sty}}$, we proceed to the answer re-annotation step, as detailed below. 

\noindent \textbf{Answer Re-annotation.} In cases where $y_{\textit{ori}}$ is incorrect for $q_{\textit{ori}}$ given $x_{\textit{sty}}$, We simply employ $\mathcal{M}$ to answer $q_{\textit{ori}}$ and generate $y_{\textit{sty}} = \mathcal{M}(P_{\textit{AR}}, x_{\textit{sty}}, q_{\textit{ori}})$.

\noindent \textbf{Question Paraphrasing.} Similar to the image captioning task, we paraphrase the given $q_{\textit{ori}}$ to address the issue of duplication. This step is more crucial in VQA tasks because the allocation of identical question phrases and answers between different images can induce shortcut learning to focus solely on the question \cite{ramakrishnan2018overcoming, agrawal2018don, jing2020overcoming, guo2021adavqa}. To address this concern, we obtain  $q_{\textit{sty}}$, a paraphrased version of $q_{\textit{ori}}$, using $q_{\textit{sty}} = \mathcal{M}(P_{\textit{QP}}, q_{\textit{ori}})$.

\subsection{Label Annotation for Visual Entailment Task}
\label{sec:annotation-label-ve}

The visual entailment task is similar to the natural language inference task. Instead of predicting the entailment of a text premise and hypothesis, the visual entailment task involves taking an image as a premise and predicting the entailment of the premise and text hypothesis. In this task, we create a data pair $d_{\textit{sty}} = (x_{\textit{sty}}, h_{\textit{sty}}, y_{\textit{sty}})$, where $h_{\textit{sty}}$ represents the hypothesis.

\noindent \textbf{Label Verification.} It is important to ensure the validity of label $y_{\textit{ori}}$ in relation to the newly generated $x_{\textit{sty}}$. To accomplish this, we use $\mathcal{M}$ to verify $y_{\textit{ori}}$ given $(x_{\textit{sty}}, h_{\textit{ori}})$, acquiring $v_{\textit{LV}} = \mathcal{M}(P_{\textit{LV}}, x_{\textit{sty}}, h_{\textit{ori}}, y_{\textit{ori}})$. If $y_{\textit{ori}}$ is not the correct label for $x_{\textit{sty}}$ and $h_{\textit{ori}}$, we proceed to label the re-annotation step as described below. Otherwise, we assign $y_{\textit{ori}}$ as $y_{\textit{sty}}$.

\noindent \textbf{Label Re-annotation.} If $y_{\textit{ori}}$ is not valid for $h_{\textit{ori}}$ given $x_{\textit{sty}}$, we utilize $\mathcal{M}$ to obtain $y_{\textit{sty}} = \mathcal{M}(P_{\textit{LR}}, x_{\textit{sty}}, h_{\textit{ori}})$.

\noindent \textbf{Hypothesis Paraphrasing.} Similar to VQA task, the use of identical hypotheses can lead to shortcut learning \cite{geirhos2020shortcut}. To address this issue, we assign the paraphrase of $h_{\textit{ori}}$ as the hypothesis $h_{\textit{sty}}$ for $x_{\textit{sty}}$. We guide $\mathcal{M}$ to paraphrase $h_{\textit{ori}}$ and obtain $h_{\textit{sty}} = \mathcal{M}(P_{\textit{HP}}, h_{\textit{ori}})$.
\section{\textsc{VolDoGer}}
\label{sec:styleshift}

Based on the annotation framework discussed in the previous section, we constructed \textsc{VolDoGer}, a dedicated dataset for domain generalization for vision-language tasks. In this section, we introduce \textsc{VolDoGer} and detail the data configuration and statistics of each dataset. In addition to the realistic photos from the original datasets, \textsc{VolDoGer} includes four distinct image styles: real photos\footnote{Note that real photos indicate the original images taken from a camera with human-annotated data.}, cartoon drawings, pencil drawings, and oil paintings. For more detailed information and analyses, please refer to Appendix~\ref{app:dataset-spec}.

\subsection{\textsc{VolDoGer-Cap}}
\textsc{VolDoGer-Cap} is a part of \textsc{VolDoGer} designed for image captioning tasks. To construct this dataset, we utilized the UIT-VIIC dataset \cite{lam2020uit}, which is a subset of the MSCOCO captioning dataset \cite{chen2015microsoft} focused on sports images, following a previous study \cite{choi2024gpts}. Consequently, \textsc{VolDoGer-Cap} contains 2695 images for training, 924 images for validation, and 231 images for testing each style. Each image is associated with five different captions.

To identify the domain gap of each style in \textsc{VolDoGer-Cap}, we use the maximum mean discrepancy (MMD) \cite{gretton2006kernel} to measure the difference in the visual and linguistic features of each domain, following previous studies \cite{zhang2021domain, chen2021mind, ren2023crossing}. Specifically, we leveraged the encoded vectors of ResNet \cite{he2016deep} and BERT \cite{devlin2019bert} to extract features from the domain and computed the MMD distances using these features. Table~\ref{tab:mmd-cap} demonstrates the result of the analysis. In this analysis, we found that \textsc{VolDoGer-Cap} exhibited a remarkable visual domain gap across every domain compared with the collection of datasets based on real photos, which was adopted by previous study \cite{ren2023crossing}, revealing the value of \textsc{VolDoGer} for domain generalization in vision-language tasks.

\begin{table}[]
\resizebox{\columnwidth}{!}{
\begin{tabular}{c|ccccc}
\Xhline{3\arrayrulewidth}
           & \textbf{R}                    & \textbf{C}                    & \textbf{P}                    & \textbf{O}              &                                     \\ \hline
\textbf{R} & -                             & {\color[HTML]{34CDF9} 0.0194} & {\color[HTML]{34CDF9} 0.0244} & {\color[HTML]{34CDF9} 0.0303} &                               \\
\textbf{C} & {\color[HTML]{FE996B} 0.0128} & -                             & {\color[HTML]{34CDF9} 0.0121} & {\color[HTML]{34CDF9} 0.0134} & \textbf{Average}              \\
\textbf{P} & {\color[HTML]{FE996B} 0.0175} & {\color[HTML]{FE996B} 0.0117} & -                             & {\color[HTML]{34CDF9} 0.0164} & {\color[HTML]{34CDF9} 0.0193} \\
\textbf{O} & {\color[HTML]{FE996B} 0.0127} & {\color[HTML]{FE996B} 0.0096} & {\color[HTML]{FE996B} 0.0111} & -                             & {\color[HTML]{FE996B} 0.0126} \\ \Xhline{3\arrayrulewidth}
\end{tabular}
}
\caption{Domain gap of each style in \textsc{VolDoGer-Cap}, measured with MMD by ResNet and BERT output vectors. Orange figures denote the visual domain gap, and blue figures represent the linguistic domain gap.}
\label{tab:mmd-cap}
\end{table}

\subsection{\textsc{VolDoGer-VQA}} \textsc{VolDoGer-VQA} is built upon the question and answer from VQA-v2 \cite{goyal2017making}, which utilizes the same images as the MSCOCO dataset and UIT-VIIC. To enhance the efficiency of the data annotation, we extracted images from the UIT-VIIC dataset along with their corresponding questions and answers. To ensure the quality and consistency of LLM-based data annotation, we exclusively used yes/no questions, as they are less ambiguous and require more direct answers than open-ended or multiple-choice questions, which can vary significantly in complexity and interpretation. Consequently, this dataset comprises 2091 images with 4120 questions for training, 711 images with 1452 questions for validation, and 182 images with 340 questions for testing for each style.

\subsection{\textsc{VolDoGer-VE}}
For \textsc{VolDoGer-VE}, we used the SNLI-VE \cite{xie2019visual} dataset, which served as the primary dataset for the visual entailment task. Similar to the approaches for \textsc{VolDoGer-Cap} and \textsc{VolDoGer-VQA}, we used only images related to football. To achieve this, we selected images containing text premise that includes the words ``soccer'' or ``football.” Subsequently, we divided them into training, validation, and test sets in a ratio of 8:1:1.  As a result, \textsc{VolDoGer-VE} comprises 619 images with 7319 hypotheses for training, 77 images with 957 hypotheses for validation, and 78 images with 856 hypotheses for testing each style.
\section{Experiment}
\label{sec:experiment}

\begin{table}[t]
\begin{center}
\resizebox{\columnwidth}{!}{
\begin{tabular}{c|ccccc}
\Xhline{3\arrayrulewidth}

\textbf{ViT} & BLEU & ROUGE & METEOR & BERTS. & BARTS.\\ \hline

\textit{Real} & 44.71 & 50.54 & 27.99 & 0.6855 & -4.6252 \\
\rowcolor{Gray}
\textit{Cartoon} &  21.29 & 33.90 & 15.10 & 0.6167 & -4.8264 \\ 
\rowcolor{Gray}
\textit{Pencil}  & 16.76 &  30.82 & 13.52 & 0.5948 &  -4.6103 \\
\rowcolor{Gray}
\textit{Oil} & 13.02 & 26.91 & 12.10 & 0.5820 & -4.7255 \\ \hline\hline
\textbf{CLIP} & BLEU & ROUGE & METEOR & BERTS. & BARTS.\\ \hline

\textit{Real}    & 41.29 & 47.76 & 25.98 & 0.6768 & -4.6252 \\ 
\rowcolor{Gray}
\textit{Cartoon} & 17.48 & 30.21 & 12.73 & 0.6021 & -4.8315 \\ 
\rowcolor{Gray}
\textit{Pencil}  & 14.07 & 27.35 & 11.15 & 0.5742 & -4.6123 \\ 
\rowcolor{Gray}
\textit{Oil}     & 11.14 & 24.31 & 10.57 & 0.5683 & -4.7301 \\ \hline\hline

\textbf{BLIP} & BLEU & ROUGE & METEOR & BERTS. & BARTS.\\ \hline

\textit{Real}    & 48.19 & 49.95 & 30.85 & 0.6950 & -4.6671 \\ 
\rowcolor{Gray}
\textit{Cartoon} & 22.65 & 32.05 & 16.12 & 0.6269 & -4.9008 \\ 
\rowcolor{Gray}
\textit{Pencil}  & 17.52 & 28.45 & 14.16 & 0.5993 & -4.7417 \\ 
\rowcolor{Gray}
\textit{Oil}     & 14.86 & 27.98 & 12.79 & 0.5868 & -4.9053 \\ 
\Xhline{3\arrayrulewidth}
\end{tabular}
}
\end{center}
\caption{Experimental results demonstrating domain shift in image captioning tasks. The \colorbox{lightgray}{highlighted} results indicate out-domain performance, reflecting models trained on \textit{Real Photos}. Please refer to Table~\ref{tab:shift-app-cap} for the results of models trained on other domains.}
\label{tab:shift-cap}
\end{table}

In this section, we describe the extensive experiments conducted using our constructed \textsc{VolDoGer} dataset and present several insights derived from the experimental results.

\begin{table}[]
\begin{center}
\resizebox{0.8\columnwidth}{!}{
\begin{tabular}{c|caaa}
\Xhline{3\arrayrulewidth}
\textbf{VQA} & \textit{Real} &\textit{ }Cartoon & \textit{Pencil} & \textit{Oil} \\ \hline
ViT                            & 55.03     & 48.52        & 47.76       & 48.65    \\ 
CLIP                           & 58.23     & 49.11        & 50.41       & 49.41    \\ 
BLIP                           & 59.19     & 50.29        & 51.32       & 50.88    \\ 
\hline\hline
\textbf{VE}  & \textit{Real}      & \textit{Cartoon}      & \textit{Pencil}      & \textit{Oil} \\ \hline
ViT       & 72.15     & 52.51        & 57.72       & 58.78    \\ 
CLIP                           & 73.10     & 55.85        & 61.61       & 60.92    \\ 
BLIP                           & 66.73     & 48.26        & 52.93       & 53.62    \\ 
\Xhline{3\arrayrulewidth}

\end{tabular}
}
\end{center}
\caption{Experimental results demonstrating domain shift in VQA and VE tasks. The \colorbox{lightgray}{highlighted} results indicate out-domain performance, reflecting models trained on \textit{Real Photos}. Please refer to Table~\ref{tab:app-vqa} and~\ref{tab:app-ve} for the results of models trained on other domains.}
\label{tab:shift-vqave}
\end{table}



\begin{table}[!ht]
  \centering
  \resizebox{\columnwidth}{!}{
  \begin{tabular}{ m{.5\textwidth} | m{.5\textwidth} }
    \Xhline{3\arrayrulewidth}
    \multicolumn{1}{c|}{\textbf{\Large In-domain Data}} & \multicolumn{1}{c}{\textbf{\Large Out-domain Data}} \\ \hline\hline

    \begin{center}
    \begin{minipage}{.3\textwidth}
    \includegraphics[width=\linewidth, height=\linewidth]{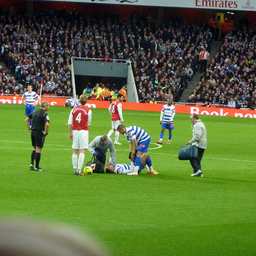}
    \end{minipage}
    \end{center}
    &
    \begin{center}
    \begin{minipage}{.3\textwidth}
    \includegraphics[width=\linewidth, height=\linewidth]{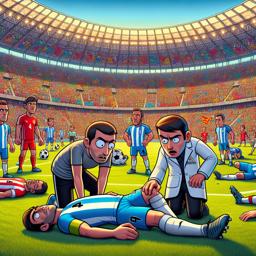}
    \end{minipage}
    \end{center}
    \\ 
    \begin{center}
    \begin{minipage}[t]{.5\textwidth}
    \Large
      \begin{itemize}
        \item A group of men playing a game of {\color[HTML]{009901} \textbf{soccer}}.
      \end{itemize}
    \end{minipage}
    \end{center}
    &
    \begin{center}
    \begin{minipage}[t]{.5\textwidth}
    \Large
      \begin{itemize}
        \item A group of people standing on {\color[HTML]{CB0000} \textbf{top of a building}}.
      \end{itemize}
    \end{minipage}
    \end{center}
    \\\\ \Xhline{2\arrayrulewidth}

    \begin{center}
    \begin{minipage}{.3\textwidth}
    \includegraphics[width=\linewidth, height=\linewidth]{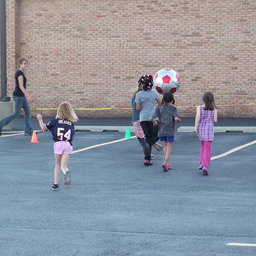}
    \end{minipage}
    \end{center}
    &
    \begin{center}
    \begin{minipage}{.3\textwidth}
    \includegraphics[width=\linewidth, height=\linewidth]{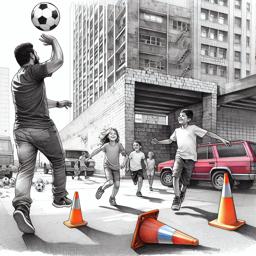}
    \end{minipage}
    \end{center}
    \\ 
    \begin{center}
    \begin{minipage}[t]{.5\textwidth}
    \Large
      \begin{itemize}
        \item A group of young {\color[HTML]{009901} \textbf{children}} playing a game of {\color[HTML]{009901} \textbf{soccer}}.
      \end{itemize}
    \end{minipage}
    \end{center}
    &
    \begin{center}
    \begin{minipage}[t]{.5\textwidth}
    \Large
      \begin{itemize}
        \item A group of young {\color[HTML]{CB0000} \textbf{men}} playing a game of {\color[HTML]{CB0000} \textbf{baseball}}.
      \end{itemize}
    \end{minipage} 
    \end{center}\\\\ \Xhline{2\arrayrulewidth}

    \begin{center}
    \begin{minipage}{.3\textwidth}
    \includegraphics[width=\linewidth, height=\linewidth]{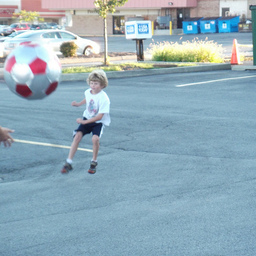}
    \end{minipage}
    \end{center}
    &
    \begin{center}
    \begin{minipage}{.3\textwidth}
    \includegraphics[width=\linewidth, height=\linewidth]{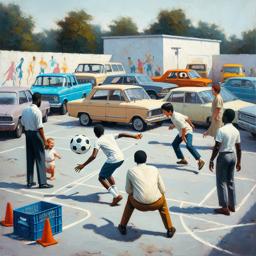}
    \end{minipage}
    \end{center}
    \\ 
    \begin{center}
    \begin{minipage}[t]{.5\textwidth}
    \Large
      \begin{itemize}
        \item A group of young men playing a game of {\color[HTML]{009901} \textbf{soccer}}.
      \end{itemize}
    \end{minipage}
    \end{center}
    &
    \begin{center}
    \begin{minipage}[t]{.5\textwidth}
    \Large
      \begin{itemize}
        \item A group of young men playing a game of {\color[HTML]{CB0000} \textbf{frisbee}}.
      \end{itemize}
    \end{minipage} 
    \end{center} \\\\ \Xhline{2\arrayrulewidth}
    
  \end{tabular}
  }
  \caption{Examples of domain shift on image captioning task. The captions are produced by a ViT-based model trained on real photos. The left side of the table showcases {\color[HTML]{009901} in-domain} examples, and the right side of the table showcases {\color[HTML]{CB0000} out-domain} examples.}
  \label{tab:shift-cap-qual}
\end{table}

\subsection{Experimental Setup}
\label{sec:experiment-setup}
First, we briefly introduce the experimental setup used in our experiments. Various models are trained using different backbones for each task. For the domain shift experiment in Section~\ref{sec:experiment-shift}, we fine-tuned the models with ViT \cite{dosovitskiy2020image} and CLIP \cite{radford2021learning} encoders with a GPT-2 \cite{radford2019language} decoder, as well as the BLIP \cite{li2022blip} model on \textsc{VolDoGer-Cap} for the image captioning task. For the VQA and VE tasks, we trained the models using the ViT and CLIP image encoders with the BERT \cite{devlin2019bert} text encoder as well as the BLIP model on \textsc{VolDoGer-VQA} and \textsc{VolDoGer-VE}. Similarly, for the domain generalization experiments in Section~\ref{sec:experiment-generalizability}, we trained the models using the ViT and frozen CLIP encoder models. In addition, we included a baseline that leveraged the ViT encoder with a dedicated technique for domain generalization, extending this approach to VQA and VE tasks \cite{ren2023crossing}.

To measure the performance of the model in the image captioning task, we employed various metrics such as BLEU \cite{papineni2002bleu}, ROUGE\_L \cite{lin2004rouge}, METEOR \cite{banerjee2005meteor}, BERTScore \cite{zhang2020bertscore}, and BARTScore \cite{yuan2021bartscore}, following previous study \cite{choi2024gpts}. The accuracy was used as a metric for the VQA and VE tasks. We report the average performance of the models trained with five random seeds. For more detailed information about the implementation of the experiment, please refer to Appendix~\ref{app:implementation-detail}.

\begin{table*}[ht!]
\begin{center}
\resizebox{\textwidth}{!}{%
\begin{tabular}{c|ccccc|c|ccccc}
\Xhline{3\arrayrulewidth}

& \multicolumn{5}{c}{\textbf{Trained on R+C+P}} & & \multicolumn{5}{c}{\textbf{Trained on R+P+O}}   \\
\textbf{ViT}                            & BLEU  & ROUGE  & METEOR  & BERTS.  & BARTS.  & \textbf{ViT} & BLEU & ROUGE  & METEOR  & BERTS.  & BARTS.  \\ \hline
\textit{Real}    & 46.46  & 51.99  & 28.79  & 0.6849  & -4.6446 & \textit{Real} & 48.42  & 53.77  & 30.23  & 0.6879  & -4.6289 \\  
\textit{Cartoon} & 42.83  & 42.07  & 23.66  & 0.6742  & -4.8284  & \cellcolor[gray]{0.9}\textit{Cartoon} & \cellcolor[gray]{0.9}32.80  & \cellcolor[gray]{0.9}36.12  & \cellcolor[gray]{0.9}20.76  & \cellcolor[gray]{0.9}0.6362  & \cellcolor[gray]{0.9}-4.8301 \\ 
\textit{Pencil}  & 42.48  & 41.16  & 23.20  & 0.6463  & -4.6108 & \textit{Pencil}  & 42.19  & 42.17  & 23.70  & 0.6541  & -4.6135  \\ 

\cellcolor[gray]{0.9}\textit{Oil}     & \cellcolor[gray]{0.9}34.07  & \cellcolor[gray]{0.9}34.75  & \cellcolor[gray]{0.9}18.71  & \cellcolor[gray]{0.9}0.6379  & \cellcolor[gray]{0.9}-4.7263 & \textit{Oil} & 45.41  & 41.73  & 23.62  & 0.6741  & -4.7233  \\ \hline\hline

\textbf{CLIP *}  & BLEU  & ROUGE  & METEOR  & BERTS.  & BARTS.  & \textbf{CLIP *} & BLEU & ROUGE  & METEOR  & BERTS.  & BARTS. \\ \hline
\textit{Real}    & 50.25  & 54.47  & 30.44  & 0.6976  & -4.6260 & \textit{Real}    & 49.41  & 54.89  & 30.94  & 0.6976  & -4.6251 \\ 
\textit{Cartoon} & 42.71  & 41.71  & 23.70  & 0.6711  & -4.8261 & \cellcolor[gray]{0.9}\textit{Cartoon} & \cellcolor[gray]{0.9}33.31  & \cellcolor[gray]{0.9}37.15  & \cellcolor[gray]{0.9}20.69  & \cellcolor[gray]{0.9}0.6337  & \cellcolor[gray]{0.9}-4.8264\\ 
\textit{Pencil}  & 42.94  & 41.59  & 23.57  & 0.6502  & -4.6120 & \textit{Pencil} & 44.32  & 43.21  & 24.87  & 0.6578  & -4.6111 \\ 

\cellcolor[gray]{0.9}\textit{Oil}     & \cellcolor[gray]{0.9}34.05  & \cellcolor[gray]{0.9}34.51  & \cellcolor[gray]{0.9}18.64  & \cellcolor[gray]{0.9}0.6366  & \cellcolor[gray]{0.9}-4.7259 & \textit{Oil} & 46.56  & 42.59  & 23.97  & 0.6774  & -4.7241 \\ \hline\hline

\textbf{ViT}$^{\dagger}$ & BLEU & ROUGE  & METEOR  & BERTS.  & BARTS.  & \textbf{ViT}$^{\dagger}$ & BLEU & ROUGE  & METEOR  & BERTS.  & BARTS. \\ \hline
\textit{Real}    & 45.49  & 51.01  & 28.04  & 0.6782  & -4.6283  & \textit{Real}  & 48.84  & 54.29  & 30.55  & 0.6945  & -4.6253   \\ 
\textit{Cartoon} & 40.55  & 40.21  & 23.12  & 0.6594  & -4.8264  & \cellcolor[gray]{0.9}\textit{Cartoon}  & \cellcolor[gray]{0.9}35.69  & \cellcolor[gray]{0.9}38.24  & \cellcolor[gray]{0.9}21.49  & \cellcolor[gray]{0.9}0.6416  & \cellcolor[gray]{0.9}-4.8253 \\ 
\textit{Pencil}  & 43.60  & 42.19  & 24.21  & 0.6516  & -4.6103  & \textit{Pencil} & 43.77  & 42.72  & 24.12  & 0.6525  & -4.6103   \\ 

\cellcolor[gray]{0.9}\textit{Oil}     & \cellcolor[gray]{0.9}36.31  & \cellcolor[gray]{0.9}36.19  & \cellcolor[gray]{0.9}19.53  & \cellcolor[gray]{0.9}0.6434  & \cellcolor[gray]{0.9}-4.7253  & \textit{Oil} & 43.27  & 40.33  & 23.46  & 0.6763  & -4.7253 \\
\Xhline{4\arrayrulewidth}

& \multicolumn{5}{c}{\textbf{Trained on R+C+O}} & & \multicolumn{5}{c}{\textbf{Trained on C+P+O}} \\
\textbf{ViT} & BLEU & ROUGE  & METEOR  & BERTS.  & BARTS.  & \textbf{ViT} & BLEU & ROUGE  & METEOR  & BERTS.  & BARTS. \\ \hline 
\textit{Real}    & 46.24  & 52.86  & 30.26  & 0.6836  & -4.6336  & \cellcolor[gray]{0.9}\textit{Real}  & \cellcolor[gray]{0.9}21.19  & \cellcolor[gray]{0.9}33.95  & \cellcolor[gray]{0.9}22.38  & \cellcolor[gray]{0.9}0.5799  & \cellcolor[gray]{0.9}-4.6305  \\ 
\textit{Cartoon} & 43.58  & 41.88  & 24.14  & 0.6755  & -4.8271  & \textit{Cartoon} & 44.05  & 43.38  & 24.47  & 0.6833  & -4.8261 \\ 

\cellcolor[gray]{0.9}\textit{Pencil}  & \cellcolor[gray]{0.9}34.18  & \cellcolor[gray]{0.9}36.55  & \cellcolor[gray]{0.9}19.77  & \cellcolor[gray]{0.9}0.6355  & \cellcolor[gray]{0.9}-4.6118  & \textit{Pencil} & 44.10  & 42.21  & 23.94  & 0.6574  & -4.6103  \\ 
\textit{Oil}     & 46.03  & 42.03  & 23.98  & 0.6752  & -4.7244  & \textit{Oil} & 47.29  & 42.50  & 23.99  & 0.6779  & -4.7253  \\ \hline\hline

\textbf{CLIP * }                   & BLEU & ROUGE  & METEOR  & BERTS.  & BARTS.  & \textbf{CLIP *} & BLEU & ROUGE  & METEOR  & BERTS.  & BARTS. \\ \hline 
\textit{Real}    & 49.02  & 53.96  & 30.73  & 0.6976  & -4.6282 & \cellcolor[gray]{0.9}\textit{Real} & \cellcolor[gray]{0.9}21.73  & \cellcolor[gray]{0.9}33.31  & \cellcolor[gray]{0.9}22.79  & \cellcolor[gray]{0.9}0.5835  & \cellcolor[gray]{0.9}-4.6252 \\ 
\textit{Cartoon} & 44.06  & 42.02  & 24.35  & 0.6829  & -4.8266 & \textit{Cartoon} & 44.48  & 43.79  & 24.33  & 0.6892  & -4.8254  \\ 

\cellcolor[gray]{0.9}\textit{Pencil}  & \cellcolor[gray]{0.9}35.40  & \cellcolor[gray]{0.9}36.89  & \cellcolor[gray]{0.9}20.19  & \cellcolor[gray]{0.9}0.6362  & \cellcolor[gray]{0.9}-4.6113 & \textit{Pencil}  & 44.65  & 42.95  & 24.46  & 0.6587  & -4.6116 \\ 
\textit{Oil}     & 47.19  & 42.77  & 23.95  & 0.6802  & -4.7231 & \textit{Oil} & 47.85  & 43.10  & 24.84  & 0.6786  & -4.7248 \\ \hline\hline

\textbf{ViT}$^{\dagger}$  & BLEU & ROUGE  & METEOR  & BERTS.  & BARTS.  & \textbf{ViT}$^{\dagger}$ & BLEU & ROUGE  & METEOR  & BERTS.  & BARTS. \\ \hline
\textit{Real}    & 47.46  & 53.09  & 30.42  & 0.6921  & -4.6253  & \cellcolor[gray]{0.9}\textit{Real}  & \cellcolor[gray]{0.9}24.51  & \cellcolor[gray]{0.9}35.06  & \cellcolor[gray]{0.9}23.63  & \cellcolor[gray]{0.9}0.6013  & \cellcolor[gray]{0.9}-4.6159  \\ 
\textit{Cartoon} & 42.49  & 41.73  & 24.41  & 0.6735  & -4.8279  & \textit{Cartoon} & 43.49  & 42.56  & 24.14  & 0.6826  & -4.8265  \\ 

\cellcolor[gray]{0.9}\textit{Pencil}  & \cellcolor[gray]{0.9}35.78  & \cellcolor[gray]{0.9}37.69  & \cellcolor[gray]{0.9}20.62  & \cellcolor[gray]{0.9}0.6461  & \cellcolor[gray]{0.9}-4.6020  & \textit{Pencil} & 43.89  & 42.07  & 23.88  & 0.6552  & -4.6119 \\ 
\textit{Oil}     & 44.86  & 41.51  & 23.86  & 0.6722  & -4.7253  & \textit{Oil}  & 46.51  & 42.03  & 23.78  & 0.6762  & -4.7254  \\ \Xhline{4\arrayrulewidth}

\end{tabular}}
\end{center}
\caption{Experimental result demonstrating the effectiveness of the domain generalization technique in image captioning tasks. * denotes that we use frozen CLIP. $\dagger$~indicates that we use ViT with \cite{ren2023crossing}.}
\label{tab:gen-cap}
\end{table*}

\begin{table*}[ht!]
\begin{center}
\resizebox{0.6\textwidth}{!}{
\begin{tabular}{c|cccc|cccc}
\Xhline{3\arrayrulewidth}
& \multicolumn{4}{c}{\textbf{VQA}} & \multicolumn{4}{c}{\textbf{VE}} \\ \hline\hline
& \multicolumn{8}{c}{\textbf{Trained on R+C+P}}  \\                                         & \textit{Real}  & \textit{Cartoon} & \textit{Pencil} & \cellcolor[gray]{0.9} \textit{Oil} & \textit{Real}  & \textit{Cartoon} & \textit{Pencil} & \cellcolor[gray]{0.9} \textit{Oil}  \\ \hline
ViT  & 55.65 & 74.11   & 75.29  & \cellcolor[gray]{0.9}75.88 & 72.20 & 71.30   & 69.11  & \cellcolor[gray]{0.9}70.04 \\ 
CLIP *                                                             & 58.52 & 76.47   & 75.58  & \cellcolor[gray]{0.9}76.59 & 72.93 & 71.63   & 69.49  & \cellcolor[gray]{0.9}70.39 \\
ViT$^{\dagger}$ & 57.35 & 73.84   & 73.52  & \cellcolor[gray]{0.9}78.82 & 71.62 & 70.73   & 68.77  & \cellcolor[gray]{0.9}70.96 \\ \hline\hline

& \multicolumn{8}{c}{\textbf{Trained on R+C+O}} \\
& \textit{Real}  & \textit{Cartoon} & \cellcolor[gray]{0.9}\textit{Pencil} & \textit{Oil} & \textit{Real}  & \textit{Cartoon} & \cellcolor[gray]{0.9}\textit{Pencil} & \textit{Oil}   \\ \hline
ViT                                                                     & 57.05 & 74.52   & \cellcolor[gray]{0.9}73.62  & 76.17 & 71.74 & 71.51   & \cellcolor[gray]{0.9}68.47  & 69.35 \\ 
CLIP *                                                             & 57.94 & 75.58   & \cellcolor[gray]{0.9}74.53  & 78.23 & 72.35 & 72.02   & \cellcolor[gray]{0.9}68.66  & 70.16 \\ 
ViT$^{\dagger}$ & 57.63 & 74.03   & \cellcolor[gray]{0.9}76.17  & 74.70 & 71.62 & 70.81   & \cellcolor[gray]{0.9}69.03  & 68.77\\ \hline\hline

& \multicolumn{8}{c}{\textbf{Trained on R+P+O}} \\                                       & \textit{Real}  & \cellcolor[gray]{0.9}\textit{Cartoon} & \textit{Pencil} & \textit{Oil} & \textit{Real}  & \cellcolor[gray]{0.9}\textit{Cartoon} & \textit{Pencil} & \textit{Oil}  \\ \hline 
ViT                                                                     & 55.58 & \cellcolor[gray]{0.9}71.76   & 76.53  & 76.94  & 71.76 & \cellcolor[gray]{0.9}67.30   & 69.22  & 69.61  \\ 
CLIP *                                                             & 56.76 & \cellcolor[gray]{0.9}73.23   & 76.85  & 77.62 & 72.11 & \cellcolor[gray]{0.9}68.19   & 70.16  & 70.09 \\ 
ViT$^{\dagger}$ & 55.29 & \cellcolor[gray]{0.9}74.41   & 76.26  & 76.79 & 71.73 & \cellcolor[gray]{0.9}69.43   & 68.95  & 69.11 \\ \hline\hline

& \multicolumn{8}{c}{\textbf{Trained on C+P+O}} \\                            & \cellcolor[gray]{0.9}\textit{Real}  & \textit{Cartoon} & \textit{Pencil} & \textit{Oil}  & \cellcolor[gray]{0.9}\textit{Real}  & \textit{Cartoon} & \textit{Pencil} & \textit{Oil}  \\ \hline 
ViT                                                                     & \cellcolor[gray]{0.9}45.53 & 74.82   & 76.62  & 77.06 & \cellcolor[gray]{0.9}58.40 & 70.65   & 69.16  & 69.54 \\ 
CLIP *                                                             & \cellcolor[gray]{0.9}47.64 & 75.18   & 77.04  & 77.15 & \cellcolor[gray]{0.9}59.31 & 71.42   & 70.21  & 70.09 \\ 
ViT$^{\dagger}$ & \cellcolor[gray]{0.9}48.82 & 74.85   & 75.94  & 76.56 & \cellcolor[gray]{0.9}59.79 & 70.20   & 68.70  & 69.23 \\ \Xhline{3\arrayrulewidth}
\end{tabular}
}
\end{center}
\caption{Experimental result demonstrating the effectiveness of domain generalization technique in VQA and VE task. * denotes that we use frozen CLIP. $\dagger$~indicates that we use ViT with \cite{ren2023crossing}. Please refer to Table~\ref{tab:app-vqa} and \ref{tab:app-ve} for the results of the models trained with other setups.}
\label{tab:gen-vqa}
\end{table*}

\subsection{Existence of Domain Shift in Vision-Language Tasks}
\label{sec:experiment-shift}

First, we investigate the existence of a domain shift using \textsc{VolDoGer}. To accomplish this, we train each model on a single domain and test it across four domains: \textbf{R}eal photo, \textbf{C}artoon drawing, \textbf{P}encil drawing, and \textbf{O}il painting. 

Tables~\ref{tab:shift-cap} and~\ref{tab:shift-vqave} list the experimental results for the three tasks. In these experiments, we observed significant differences between the in-domain and out-domain performances, confirming the existence of a domain shift in response to input images with different styles. Examples of the outputs produced by a captioning model solely using real photos in Table~\ref{tab:shift-cap-qual} support the experimental results. In these examples, we can observe that the model cannot accurately generate descriptions for images with similar content but different styles. While this phenomenon has been observed in other tasks, such as image classification \cite{peng2019moment}, validating its existence in vision-language tasks has been challenging because of the absence of a dedicated dataset. Our study demonstrates that this phenomenon persists in vision-language tasks using \textsc{VolDoGer}, underscoring the need for future research in this area.

\subsection{Effectiveness of Domain Generalization Techniques to Mitigate Domain Shift}
\label{sec:experiment-generalizability}

Subsequently, we evaluate the effectiveness of the domain generalization techniques in mitigating the domain shift identified, as discussed in the previous section. In this experiment, we employed the domain generalization method from a previous study using a ViT encoder \cite{ren2023crossing}. Because this strategy focused solely on the image captioning task, we extended it to VQA and VE tasks, with the modifications detailed in Appendix~\ref{app:implementation-detail}. We established two baselines for this experiment: joint training without a dedicated strategy using ViT encoders and fixed CLIP encoders.

Tables~\ref{tab:gen-cap},~\ref{tab:gen-vqa} list the experimental results. In general, we found that using multiple source domains enhanced the out-domain performance, as indicated in red in the tables, compared to models trained on a single domain. Additionally, we discovered that the dedicated domain generalization strategy for vision-language tasks is more beneficial for out-domain performance than naive joint training. However, the implementation of such a strategy exhibited a slightly lower in-domain performance than the baselines. This highlights the potential for improvements in domain generalization techniques for vision-language tasks. We believe that the proposed \textsc{VolDoGer} will play a crucial role in the development and benchmarking of this direction.

\section{Conclusion}
\label{sec:conclusion}

In this study, we propose a data annotation framework that leverages multimodal LLMs to construct a dataset with various styles for vision-language tasks. We created \textsc{VolDoGer}, a dataset for three vision-language tasks with four different image styles by exploiting the proposed pipeline. Using \textsc{VolDoGer}, we conducted extensive experiments across three tasks using various models. Our experiments confirmed the existence of a domain shift in vision-language tasks when dealing with images in different styles compared with the training data. In addition, we validated the effectiveness of the domain generalization strategy in our setup. We believe that our framework and \textsc{VolDoGer} will serve as cornerstones for future research on domain generalization for vision-language tasks.
\section{Discussion}
\label{sec:discussion}

In this section, we discuss the potential limitations of our study. First, it should be noted that the primary consideration of \textsc{VolDoGer} is the stylistic domain shift of image in vision-language tasks, rather than other elements such as cultural representation of the image or linguistic differences. Nevertheless, our \textsc{VolDoGer} is a dedicated dataset for evaluating and mitigating stylistic domain shift, playing a complementary role with the dataset for semantic domain shift proposed by previous study \cite{ren2023crossing}. 

Second, the analysis presented in Appendix~\ref{app:dataset-spec} regarding the distribution of labels in each dataset, as depicted in Figure~\ref{fig:spec-vqa} and~\ref{fig:spec-ve}, revealed that the distribution of the label differs from that of the original VQA and VE datasets. This is attributed to the difference between $x_{\textit{ori}}$ and $x_{\textit{sty}}$, which is marginal in general, but can alter the label of the question or hypothesis. For instance, the example in Appendix~\ref{app:additional-examples-annotation-vqa} shows a change in the label regarding the question. In particular, the answer to the question asking the position of the tennis athletes is ``Yes'' for $x_{\textit{ori}}$ but ``No'' for $x_{\textit{sty}}$. We acknowledge that this difference between $x_{\textit{ori}}$ and $x_{\textit{sty}}$ arises because of our use of generative models. Additionally, the difference between captions could raise the trade-off between paraphrasing and parallelism, which is important for domain generalization. However, it should be considered that completely parallel data in different domains or styles can be difficult to gather in real-world scenarios. Accordingly, the current setup with slightly different lexical expression could better reflect the real-world scenario, while maintaining the core semantics of the images.

Such differences between $x_{\textit{ori}}$ and $x_{\textit{sty}}$ could be minimized with the adoption of other methods including AdaIN \cite{huang2017arbitrary} or StyleGAN \cite{karras2019style}. However, these methods could induce another type of spurious correlation of models that overly relies on the texture or shape of the image \cite{geirhos2019imagenet, naseer2021intriguing}. In the meantime, more meticulous verification methods for generated images, such as TIFA \cite{hu2023tifa} could potentially mitigate this issue by ensuring the similarity between $x_{\textit{ori}}$ and $x_{\textit{sty}}$, such restriction could raise the cost for overall progress. Considering this, in future work, we will focus on improving the proposed annotation method such that it considers the preservation of the label and maintains label distribution, with the consideration of the trade-off between cost and correctness.

{
    \small
    \bibliographystyle{ieeenat_fullname}
    \bibliography{main}
}

\clearpage
\setcounter{page}{1}
\maketitlesupplementary

\section{Evaluation on Zero-shot Performance of Multimodal Large Language Models}
\label{sec:experiment-zeroshot}
In this section, we present the evaluation of the zero-shot performance of recent multimodal LLMs that can perform various tasks without specific training \cite{yin2023survey}. We adopted open-source models such as BLIP-2 \cite{li2023blip}, PaliGemma \cite{google2024paligemma}, and LLaVA \cite{liu2023visual}, as well as proprietary models such as GPT-4 \cite{openai2023gpt}, Gemini \cite{google2024gemini}, and Claude 3 \cite{anthropic2024claude} for a comprehensive evaluation.

The results are listed in Tables~\ref{tab:zeroshot-cap},~\ref{tab:zeroshot-vqa}, and~\ref{tab:zeroshot-ve}. Overall, GPT-4o demonstrated the best performance in most cases. Additionally, we observed that several open-source models outperformed proprietary models in VQA tasks with real images but not in other tasks such as VE and image captioning. Notably, PaliGemma and LLaVA 1.5 exhibited considerably worse performance than the other models. This phenomenon may indicate the possibility of task contamination \cite{li2024task}, where these open-source models may have used VQA-v2 data during their training process\footnote{Note that PaliGemma clarified that they used a mixture of downstream academic datasets.}.

The possibility of task contamination suggests that our proposed \textsc{VolDoGer} may not be optimal for measuring the zero-shot performance of multimodal LLMs. While we identified potential task contamination based on the performance discrepancies between PaliGemma and LLaVA 1.5 models on the VQA and VE tasks, other models, including proprietary models, may also exploit the original datasets, such as VQA-v2, SNLI-VE, and MSCOCO images, which we utilized to construct \textsc{VolDoGer}.

This underscores the need for a more sophisticated approach for comparing the zero-shot performance of multimodal LLMs across different styles. One potential strategy for addressing this gap is to compare the outputs produced by various models based on human preferences \cite{chiang2024chatbot}. Specifically, this could involve crowdsourcing the collection of human preferences for different models based on specific images and ranking the models using these data.

Despite its limitations in measuring the zero-shot performance of multimodal LLMs, \textsc{VolDoGer} is the first dedicated dataset for domain generalization across multiple vision-language tasks with different styles. This will serve as a valuable resource for future research on domain generalization for these tasks.

\begin{table}[]
\begin{center}
\resizebox{\columnwidth}{!}{
\begin{tabular}{c|ccccc}
\Xhline{3\arrayrulewidth}
\multirow{2}{*}{\textbf{Captioning}}                               & \multicolumn{4}{c}{\textbf{Open-Source Models}} \\
                                                                   & Real   & Cartoon & Pencil   & Oil               \\ \hline\hline
BLIP2-FlanT5-XL                                                    & -6.395 & -6.822  & -6.516   & -6.693            \\ \hline
PaliGemma                                                          & -4.754 & -5.868  & -5.114   & -5.091            \\ \hline
LLaVA 1.5                                                          & -4.625 & -4.829  & -4.618   & -4.725             \\ \hline
\begin{tabular}[c]{@{}c@{}}LLaVA-NeXT\\ w/ Vicuna-7B\end{tabular}  & -4.652 & -4.883  & -4.644   & -4.724             \\ \hline
\begin{tabular}[c]{@{}c@{}}LLaVA-NeXT\\ w/ Mistral-7B\end{tabular} & -4.698 & -5.023  & -4.702   & -4.846            \\ \Xhline{2\arrayrulewidth}
\multirow{2}{*}{\textbf{Captioning}}                               & \multicolumn{4}{c}{\textbf{Proprietary Models}} \\
                                                                   & Real   & Cartoon & Pencil   & Oil               \\ \hline\hline
\begin{tabular}[c]{@{}c@{}}GPT-4-Vision\\ 1106-preview\end{tabular}& -4.629 & -4.827  & -4.618   & -4.725             \\ \hline
\begin{tabular}[c]{@{}c@{}}GPT-4-Turbo\\ 2024-04-09\end{tabular}   & -4.625 & -4.829  & -4.619   & -4.725            \\ \hline
\begin{tabular}[c]{@{}c@{}}GPT-4o\\ 2024-05-13\end{tabular}        & -4.636 & -4.836  & -4.623   & -4.726             \\ \hline
Claude 3 Haiku                                                     & -4.640 & -4.829  & -4.624   & -4.726             \\ \hline
Claude 3 Sonnet                                                    & -4.630 & -4.828  & -4.617   & -4.726            \\ \hline
Claude 3 Opus                                                      & -4.639 & -4.829  & -4.620   & -4.726            \\ \hline
Gemini 1.0 Pro                                                     & -4.626 & -4.829  & -4.611   & -4.725            \\ \hline
Gemini 1.5 Flash                                                   & -4.725 & -4.618  & -4.827   & -4.625            \\ \Xhline{3\arrayrulewidth}
\end{tabular}
}
\end{center}
\caption{Experimental result demonstrating the zero-shot performance of multimodal LLMs on image captioning task. We only report BARTScore for this experiment as matching-based metrics are less suitable for evaluating the quality of zero-shot text generation \cite{fu2023gptscore}.}
\label{tab:zeroshot-cap}
\end{table}
\begin{table}[]
\begin{center}
\resizebox{\columnwidth}{!}{
\begin{tabular}{c|ccccc}
\Xhline{3\arrayrulewidth}
\multirow{2}{*}{\textbf{VQA}}                                      & \multicolumn{4}{c}{\textbf{Open-Source Models}} \\
                                                                   & Real   & Cartoon & Pencil   & Oil               \\ \hline\hline
BLIP2-FlanT5-XL                                                    & 65.29  & 64.41   & 61.18    & 62.92             \\ \hline
PaliGemma                                                          & 80.59  & 79.41   & 75.29    & 75.59             \\ \hline
LLaVA 1.5                                                          & 80.88  & 76.18   & 72.94    & 71.18             \\ \hline
\begin{tabular}[c]{@{}c@{}}LLaVA-NeXT\\ w/ Vicuna-7B\end{tabular}  & 80.29  & 67.65   & 64.12    & 64.12             \\ \hline
\begin{tabular}[c]{@{}c@{}}LLaVA-NeXT\\ w/ Mistral-7B\end{tabular} & 81.76  & 65.88   & 61.18    & 64.41             \\ \Xhline{2\arrayrulewidth}
\multirow{2}{*}{\textbf{VQA}}                                      & \multicolumn{4}{c}{\textbf{Proprietary Models}} \\
                                                                   & Real   & Cartoon & Pencil   & Oil               \\ \hline\hline
\begin{tabular}[c]{@{}c@{}}GPT-4-Vision\\ 1106-preview\end{tabular}& 75.29  & 67.06   & 59.12    & 62.35             \\ \hline
\begin{tabular}[c]{@{}c@{}}GPT-4-Turbo\\ 2024-04-09\end{tabular}   & 76.47  & 67.65   & 62.94    & 64.71             \\ \hline
\begin{tabular}[c]{@{}c@{}}GPT-4o\\ 2024-05-13\end{tabular}        & 77.35  & 82.94   & 79.41    & 78.53             \\ \hline
Claude 3 Haiku                                                     & 75.00  & 67.35   & 62.06    & 62.35             \\ \hline
Claude 3 Sonnet                                                    & 68.24  & 74.12   & 72.35    & 70.29             \\ \hline
Claude 3 Opus                                                      & 63.53  & 63.82   & 61.76    & 63.24             \\ \hline
Gemini 1.0 Pro                                                     & 73.23  & 68.24   & 68.23    & 68.82             \\ \hline
Gemini 1.5 Flash                                                   & 75.88  & 78.82   & 73.82    & 72.94             \\ \Xhline{3\arrayrulewidth}
\end{tabular}
}
\end{center}
\caption{Experimental result demonstrating the zero-shot performance of multimodal large language models in VQA task.}
\label{tab:zeroshot-vqa}
\end{table}
\begin{table}[]
\begin{center}
\resizebox{\columnwidth}{!}{
\begin{tabular}{c|ccccc}
\Xhline{3\arrayrulewidth}
\multirow{2}{*}{\textbf{VE}}                                       & \multicolumn{4}{c}{\textbf{Open-Source Models}} \\
                                                                   & Real   & Cartoon & Pencil   & Oil               \\ \hline\hline
BLIP2-FlanT5-XL                                                    & 63.82  & 73.13   & 72.24    & 72.00             \\ \hline
PaliGemma                                                          & 34.33  & 33.91   & 35.02    & 34.79             \\ \hline
LLaVA 1.5                                                          & 33.53  & 29.87   & 33.41    & 32.60             \\ \hline
\begin{tabular}[c]{@{}c@{}}LLaVA-NeXT\\ w/ Vicuna-7B\end{tabular}  & 55.76  & 55.25   & 57.95    & 55.18             \\ \hline
\begin{tabular}[c]{@{}c@{}}LLaVA-NeXT\\ w/ Mistral-7B\end{tabular} & 70.05  & 70.36   & 67.86    & 69.24             \\ \Xhline{2\arrayrulewidth}
\multirow{2}{*}{\textbf{VE}}                                       & \multicolumn{4}{c}{\textbf{Proprietary Models}} \\
                                                                   & Real   & Cartoon & Pencil   & Oil               \\ \hline\hline
\begin{tabular}[c]{@{}c@{}}GPT-4-Vision\\ 1106-preview\end{tabular}& 65.32  & 70.59   & 70.51    & 71.20             \\ \hline
\begin{tabular}[c]{@{}c@{}}GPT-4-Turbo\\ 2024-04-09\end{tabular}   & 61.75  & 72.43   & 72.58    & 70.05             \\ \hline
\begin{tabular}[c]{@{}c@{}}GPT-4o\\ 2024-05-13\end{tabular}        & 71.08  & 73.13   & 72.47    & 70.74             \\ \hline
Claude 3 Haiku                                                     & 58.18  & 63.55   & 67.86    & 66.47             \\ \hline
Claude 3 Sonnet                                                    & 59.22  & 72.78   & 72.24    & 71.08             \\ \hline
Claude 3 Opus                                                      & 59.91  & 66.65   & 61.18    & 64.06             \\ \hline
Gemini 1.0 Pro                                                     & 64.63  & 60.32   & 63.13    & 64.29             \\ \hline
Gemini 1.5 Flash                                                   & 64.17  & 74.39   & 73.96    & 72.35             \\ \Xhline{3\arrayrulewidth}
\end{tabular}
}
\end{center}
\caption{Experimental result demonstrating the zero-shot performance of multimodal large language models in VE task.}
\label{tab:zeroshot-ve}
\end{table}

\section{Implementation Detail}
\label{app:implementation-detail}
This section presents implementation details to supplement the experimental setup described in Section~\ref{sec:experiment-setup}. We primarily employed PyTorch \cite{paszke2019pytorch} and Transformers \cite{wolf2020transformers} to this end. The training and inference of the fine-tuned models were performed on a single Nvidia RTX 3090 GPU, whereas the inference of multimodal large language models was conducted on a single Nvidia A100 GPU. Please refer to the source code for the annotated data and more details\footnote{\url{https://anonymous.4open.science/r/VL_LLM_ANNO}}.

\subsection{Image Captioning}
\noindent \textbf{Fine-tuned Models.} For image captioning, we used ViT and CLIP encoders with a GPT-2 decoder. Specifically, we adopted \textit{google/vit-base-patch16-224-in21k}, \textit{openai/clip-vit-base-patch16}, and \textit{openai-community/gpt2} from Transformers library, respectively. For BLIP, we avoided directly applying \textit{Salesforce/blip-image-captioning-base} as our baseline because this model had already used the MSCOCO captioning dataset for continual pre-training. Instead, we loaded the raw checkpoint of the BLIP model before pre-training. Every model was trained based on the Adam \cite{kingma2015adam} optimizer with a learning rate of 5e-5 for three epochs without the deployment of a scheduler. The batch size of the model was set to 16. Each input image was resized to 256$\times$256 size and the region with 224$\times$224 size was randomly cropped from the resized image during training. For inference, a 224$\times$224 region was obtained from the center of the resized image. This resizing and cropping procedure was applied to each model and across all three tasks.

\noindent \textbf{Domain Generalization Method.} We implemented the domain generalization method we used for our experiment from scratch because there is no available source code \cite{ren2023crossing}. Although we followed their explanation to implement the framework, it is important to note that we used the encoded feature of the ViT encoder instead of the ResNet \cite{he2016deep} model.

\noindent \textbf{Zero-shot Models.} We adopted \textit{google/paligemma-3b-mix-224}, \textit{llava-hf/llava-1.5-7b-hf}, \textit{llava-hf/llava-v1.6-vicuna-7b-hf}, and \textit{llava-hf/llava-v1.6-mistral-7b-hf} from Transformers as PaliGemma, LLaVA-1.5, LLaVA-NeXT w/ Vicuna, and LLaVA-NeXT w/ Mistral, respectively, in our experiments. We used slightly different input prompts for the open-source and proprietary models because proprietary models offer system prompts. For open-source models, we used a relatively simple prompt, which is ``Provide a detailed description of the given image in one sentence.'' For proprietary models involving GPT-4, Claude, and Gemini, we applied the following system prompt: ``You are a helpful AI assistant that helps people generate captions for their images. Your output should be a single sentence that describes the image. Do not generate any inappropriate or accompanying text.'' The input prompt was set to ``Please generate a caption for this image. Please generate the result in the form of Caption: <your caption here>''.

\noindent \textbf{Evaluation Metric.} The NLG-EVAL library \cite{sharma2017nlgeval} was used to measure the BLEU, ROUGE, and METEOR metrics. We reported the average of BLEU-1, 2, 3, and 4 scores as BLEU score. For BERTScore and BARTScore, we adopted the \textit{bert-base-uncased} and \textit{facebook/bart-large-cnn} models, respectively.

\subsection{Visual Question Answering}
\noindent \textbf{Fine-tuned Models.} We used identical models for ViT and CLIP image encoders. For the BERT text encoder, we adopted the \textit{bert-base-uncased} model. Each output feature produced by the image and text encoders with a vector size of 768 was concatenated into a single feature with a size of 1536, and was fed into the classifier with a single ReLU activation. For the BLIP model, we used the raw checkpoint instead of the \textit{Salesforce/blip-vqa-base}. We trained each model with a learning rate of 5e-5 for 10 epochs using the Adam optimizer, with early stopping based on the accuracy of the validation set.

\noindent \textbf{Domain Generalization Method.} We used the image caption offered by the MSCOCO captioning dataset because VQA-v2 dataset was also built on images from MSCOCO.

\noindent \textbf{Zero-shot Models.} For open-source models that do not support a dedicated system prompt, we used the following simple prompt: ``Question: based on the image, \{question\}? Answer with yes or no.'' For proprietary models, we applied the following system prompt: ``You are a helpful AI assistant that helps visual question answering tasks.'', while the input prompt was set to ``Please answer the question below based on the given image. Start the response with Yes or No. Question: \{question\}?'' This choice was made as open-source models such as PaliGemma do not support system prompts.

\subsection{Visual Entailment}
\noindent \textbf{Fine-tuned Models.} We used a model structure identical to that used for VQA task. The models were trained using the Adam optimizer for three epochs with a learning rate of 5e-5.

\noindent \textbf{Domain Generalization Method.} We used the text premise offered by SNLI-VE dataset as a description of a given image, as they are the captions from Flickr30k dataset, the source of the image of SNLI-VE.

\noindent \textbf{Zero-shot Models.} We used the following simple prompt for open-source models: ``Statement: \{hypothesis\} Determine if the statement is true, false, or undetermined based on the image. Answer with true, false, or undetermined.'' For proprietary models, we applied the following system prompt: ``You are a helpful AI assistant that helps visual entailment tasks.'', and the input prompt applied was ``Does the given hypothesis entail the image? Start the response with True, False, or Undetermined. Hypothesis: \{hypothesis\}''

\subsection{Data Annotation}
For data annotation, we used the GPT-4o \cite{openai2024gpt4o} model as our $\mathcal{M}$. The model version was \textit{GPT-4o-2024-05-13}. We set every parameter, including the top-p and temperature as default. We set the patience of error to 10, and the data that exceeded this patience were omitted from the annotation procedure. Prompts for the annotation process such as $P_{\textit{ID}}$ are provided in Appendix~\ref{app:prompt-data-annotation}. In addition, we used DALL-E 3 \cite{betker2023improving} as the image generation model $\mathcal{G}$. Note that other image generation models such as Stable Diffusion \cite{esser2024scaling} can also be used as $\mathcal{G}$ instead of DALL-E 3. The overall data annotation procedure costs approximately USD 1,800.

\section{Dataset Specification}
\label{app:dataset-spec}
\begin{figure*}[]
    \centering
    \includegraphics[width=0.8\textwidth]{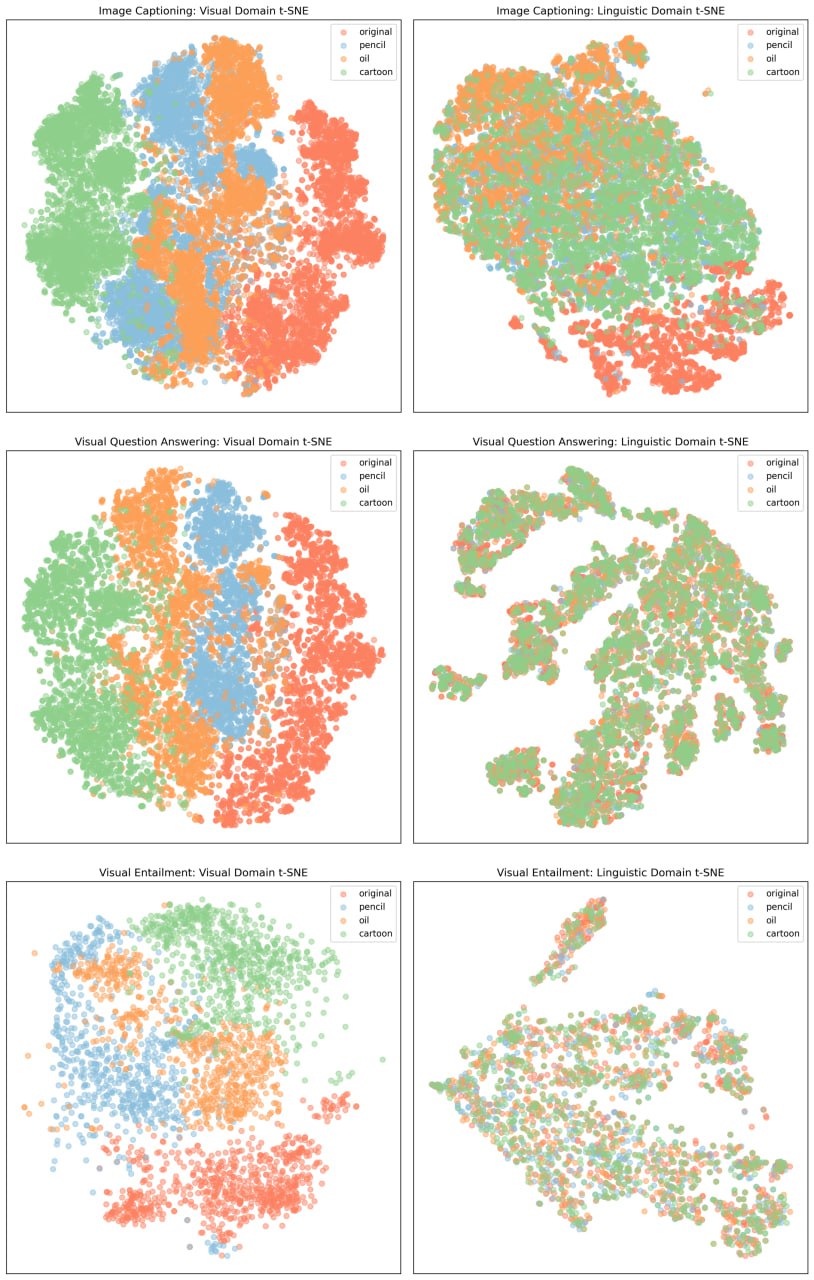}
    \caption{The t-SNE visualization result of each domain on three tasks.}
\label{fig:tsne-total}
\end{figure*}

In this section, we provide more detailed information on the \textsc{VolDoGer}. Additionally, Figure~\ref{fig:tsne-total} suggests the result t-SNE visualization \cite{van2008visualizing} for each domain of three tasks, especially demonstrating visual domain gaps.

\subsection{\textsc{VolDoGer-Cap}}

Table~\ref{tab:spec-cap} lists the number of images for each style in \textsc{VolDoGer-Cap}. Each style contains approximately 3,850 images, with five different captions for each image.

\begin{table}[h]
\begin{center}
\resizebox{\columnwidth}{!}{
\begin{tabular}{c|cccc}
\Xhline{3\arrayrulewidth}
\textbf{Captioning} & Train & Validation & Test & Total \\ \hline\hline
Real                & 2695  & 924        & 231  & 3850  \\ \hline
Cartoon             & 2695  & 924        & 231  & 3850  \\ \hline
Pencil              & 2694  & 923        & 231  & 3848  \\ \hline
Oil                 & 2694  & 924        & 231  & 3849  \\ \Xhline{3\arrayrulewidth}
\end{tabular}
}
\end{center}
\caption{The amount of images for each style in \textsc{VolDoGer-Cap}.}
\label{tab:spec-cap}
\end{table}

\subsection{\textsc{VolDoGer-VQA}}

\begin{figure*}[]
    \centering
    \includegraphics[width=\textwidth]{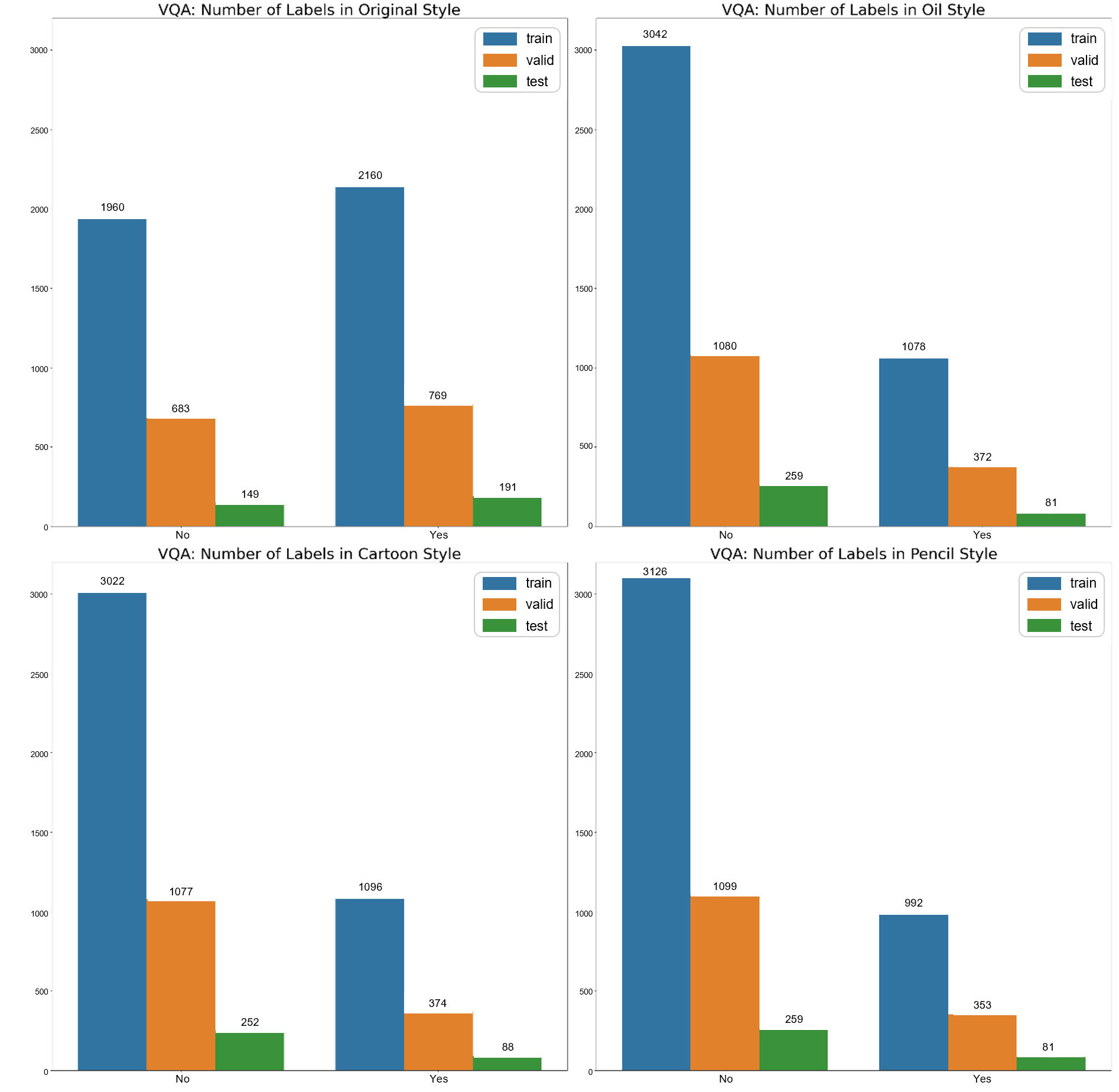}
    \caption{The label distribution of \textsc{VolDoGer-VQA} for each split.}
\label{fig:spec-vqa}
\end{figure*}

Tables~\ref{tab:spec-vqa} and~\ref{tab:mmd-vqa} present the number of images and questions as well as the domain gap for each style in \textsc{VolDoGer-VQA}. Figure~\ref{fig:spec-vqa} presents the number of labels for each split.

\begin{table}[h]
\begin{center}
\resizebox{\columnwidth}{!}{
\begin{tabular}{c|cccc}
\Xhline{3\arrayrulewidth}
\textbf{\begin{tabular}[c]{@{}c@{}}VQA\\ Images\end{tabular}}    & Train & Valid & Test & Total \\ \hline\hline
Real                                                             & 2091  & 711   & 182  & 2984  \\ \hline
Cartoon                                                          & 2090  & 710   & 182  & 2982  \\ \hline
Pencil                                                           & 2090  & 711   & 182  & 2983  \\ \hline
Oil                                                              & 2091  & 711   & 182  & 2984  \\ \Xhline{2\arrayrulewidth}
\textbf{\begin{tabular}[c]{@{}c@{}}VQA\\ Questions\end{tabular}} & Train & Valid & Test & Total \\ \hline\hline
Real                                                             & 4120  & 1452  & 340  & 5912  \\ \hline
Cartoon                                                          & 4118  & 1451  & 340  & 5909  \\ \hline
Pencil                                                           & 4118  & 1452  & 340  & 5910  \\ \hline
Oil                                                              & 4120  & 1452  & 340  & 5912  \\ \Xhline{3\arrayrulewidth}
\end{tabular}
}
\end{center}
\caption{The number of images and questions for each style in \textsc{VolDoGer-VQA}.}
\label{tab:spec-vqa}
\end{table}

\begin{table}[]
\resizebox{\columnwidth}{!}{
\begin{tabular}{c|ccccc}
\Xhline{3\arrayrulewidth}
           & \textbf{R}                    & \textbf{C}                    & \textbf{P}                    & \textbf{O}              &                                     \\ \hline
\textbf{R} & -                             & {\color[HTML]{34CDF9} 0.0024} & {\color[HTML]{34CDF9} 0.0026} & {\color[HTML]{34CDF9} 0.0026} &                               \\
\textbf{C} & {\color[HTML]{FE996B} 0.0127} & -                             & {\color[HTML]{34CDF9} 0.0016} & {\color[HTML]{34CDF9} 0.0016} & \textbf{Average}              \\
\textbf{P} & {\color[HTML]{FE996B} 0.0165} & {\color[HTML]{FE996B} 0.0109} & -                             & {\color[HTML]{34CDF9} 0.0014} & {\color[HTML]{34CDF9} 0.0020} \\
\textbf{O} & {\color[HTML]{FE996B} 0.0124} & {\color[HTML]{FE996B} 0.0091} & {\color[HTML]{FE996B} 0.0106} & -                             & {\color[HTML]{FE996B} 0.0120} \\ \Xhline{3\arrayrulewidth}
\end{tabular}
}
\caption{Domain gap of each style in \textsc{VolDoGer-VQA}, measured with MMD by ResNet and BERT output vectors. Orange figures denote the visual domain gap, and blue figures represent the linguistic domain gap.}
\label{tab:mmd-vqa}
\end{table}

\subsection{\textsc{VolDoGer-VE}}

\begin{figure*}[]
    \centering
    \includegraphics[width=\textwidth]{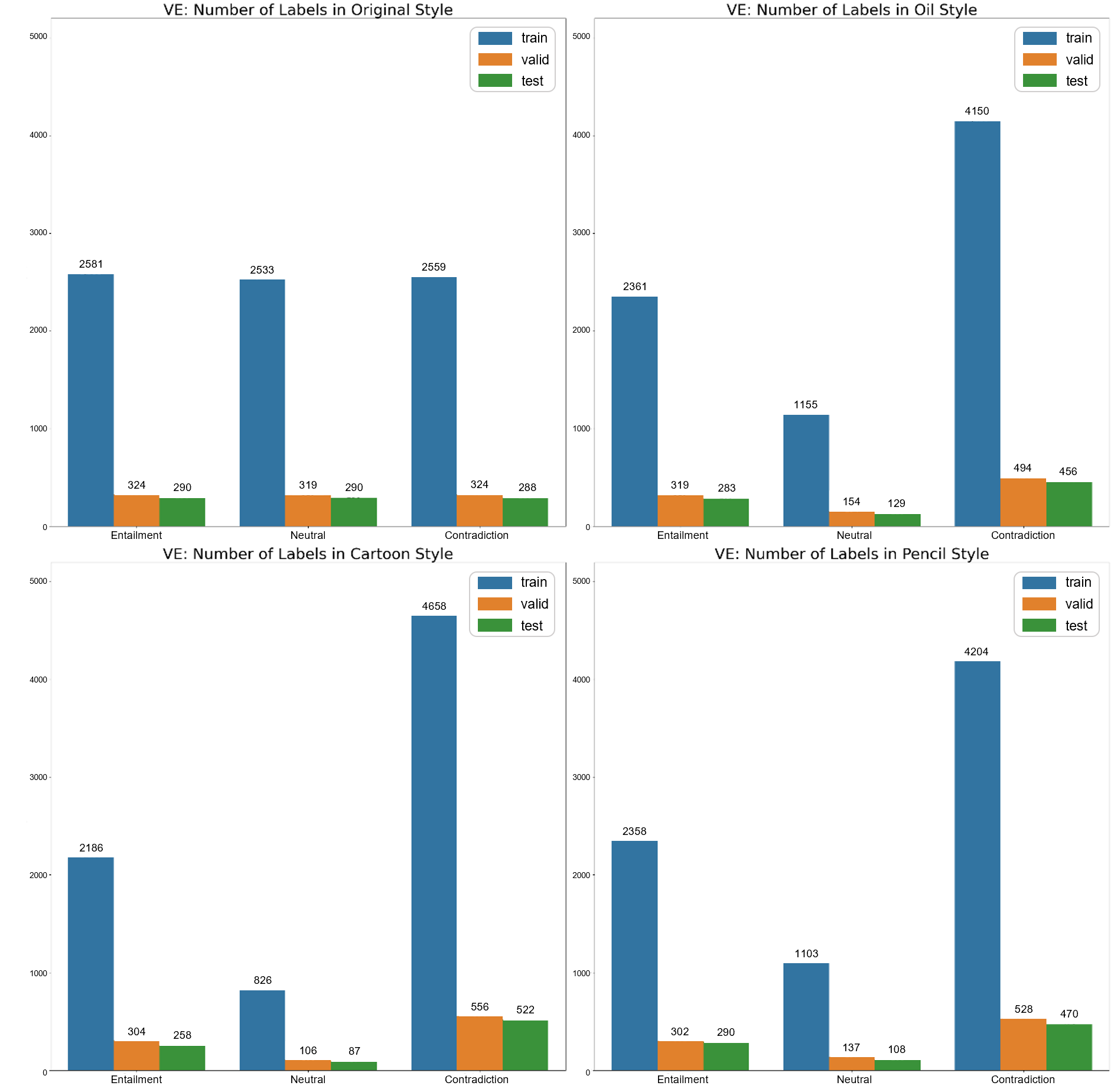}
    \caption{The label distribution of \textsc{VolDoGer-VE} for each split.}
\label{fig:spec-ve}
\end{figure*}

Tables ~\ref{tab:spec-ve} and~\ref{tab:mmd-ve} present the number of images, hypotheses, and the domain gap for each style in \textsc{VolDoGer-VE}. Figure~\ref{fig:spec-ve} presents the number of labels for each split.

\begin{table}[h]
\begin{center}
\resizebox{\columnwidth}{!}{
\begin{tabular}{c|cccc}
\Xhline{3\arrayrulewidth}
\textbf{\begin{tabular}[c]{@{}c@{}}VE\\ Images\end{tabular}}     & Train & Valid & Test & Total \\ \hline\hline
Real                                                             & 619   & 77    & 78   & 774   \\ \hline
Cartoon                                                          & 618   & 77    & 78   & 773   \\ \hline
Pencil                                                           & 619   & 77    & 78   & 774   \\ \hline
Oil                                                              & 619   & 77    & 78   & 774   \\ \Xhline{2\arrayrulewidth}
\textbf{\begin{tabular}[c]{@{}c@{}}VE\\ Hypotheses\end{tabular}} & Train & Valid & Test & Total \\ \hline\hline
Real                                                             & 7673  & 967   & 868  & 9508  \\ \hline
Cartoon                                                          & 7670  & 966   & 867  & 9503  \\ \hline
Pencil                                                           & 7665  & 967   & 868  & 9500  \\ \hline
Oil                                                              & 7666  & 967   & 868  & 9501  \\ \Xhline{3\arrayrulewidth}
\end{tabular}
}
\end{center}
\caption{Number of images and questions for each style in \textsc{VolDoGer-VE}.}
\label{tab:spec-ve}
\end{table}

\begin{table}[]
\resizebox{\columnwidth}{!}{
\begin{tabular}{c|ccccc}
\Xhline{3\arrayrulewidth}
           & \textbf{R}                    & \textbf{C}                    & \textbf{P}                    & \textbf{O}              &                                     \\ \hline
\textbf{R} & -                             & {\color[HTML]{34CDF9} 0.0060} & {\color[HTML]{34CDF9} 0.0067} & {\color[HTML]{34CDF9} 0.0062} &                               \\
\textbf{C} & {\color[HTML]{FE996B} 0.0109} & -                             & {\color[HTML]{34CDF9} 0.0042} & {\color[HTML]{34CDF9} 0.0044} & \textbf{Average}              \\
\textbf{P} & {\color[HTML]{FE996B} 0.0146} & {\color[HTML]{FE996B} 0.0109} & -                             & {\color[HTML]{34CDF9} 0.0038} & {\color[HTML]{34CDF9} 0.0052} \\
\textbf{O} & {\color[HTML]{FE996B} 0.0106} & {\color[HTML]{FE996B} 0.0087} & {\color[HTML]{FE996B} 0.0104} & -                             & {\color[HTML]{FE996B} 0.0110} \\ \Xhline{3\arrayrulewidth}
\end{tabular}
}
\caption{Domain gap of each style in \textsc{VolDoGer-VE}, measured with MMD by ResNet and BERT output vectors. Orange figures denote the visual domain gap, and blue figures represent the linguistic domain gap.}
\label{tab:mmd-ve}
\end{table}

\section{Ablation Study}
\label{app:ablation-study}

In this section, we conduct an ablation study that validates the effectiveness of label verification and re-annotation in VQA and VE tasks.

\subsection{Manual Analysis on Label Verification}
First, we manually investigated the results of label verification and label re-annotation. We selected the test split of three styles in the VQA task as representatives. Subsequently, we gathered data with labels that differed from those in the real photo domain. As a result, we acquired 127 questions from the cartoon drawing domain, 134 questions from the pencil drawing domain, and 130 questions from the oil painting domain. We then examined the annotation results to determine their acceptability. We found that 26 questions from the cartoon drawing domain, 24 questions from the pencil drawing domain, and 25 questions from the oil painting domain were unacceptable and falsely annotated, accounting for less than 20\% of each domain.

Furthermore, we observed several tendencies in LLM-based annotations. For instance, the LLM predominantly suggested ``No'' for subjective questions such as ``Is the weather cold?'', ``Is this man happy?'', or ``Is the boy good at this game?''. Moreover, the LLM struggled with questions asking about the professionalism of a game, such as ``Is this a major league game?''. We aim to investigate these tendencies more thoroughly in future work. Additionally, This analysis is included in the dataset repository as a report, providing a broad perspective and assisting future studies.

\subsection{Experiment on Answer Verification}

Second, we conducted an ablation experiment by directly assigning labels from the real photo domain, thereby excluding the answer verification process. We created an ablation training set based on this setup and trained three VQA models for each style, evaluating their performance on in-domain test sets. The results are presented in Table~\ref{tab:app-ablation}. The findings suggest that directly assigning labels from the real photo domain to other domains can harm model performance, as the distinction between real and generated images, along with their labels, acts as noisy labels.

In conclusion, both the manual analysis and the experimental results support the significance of the answer and label verification and re-annotation procedure we proposed in Section~\ref{sec:annotation-label-vqa} and ~\ref{sec:annotation-label-ve}.

\begin{table}[]
\begin{center}
\resizebox{\columnwidth}{!}{
\begin{tabular}{c|ccc}
\Xhline{3\arrayrulewidth}
\textbf{VQA}                       & Cartoon & Pencil & Oil   \\ \hline\hline
w/ Answer Verification (Ours)      & 75.23   & 75.29  & 77.35 \\
w/o Answer Verification (Ablation) & 71.17   & 73.23  & 75.58 \\ \Xhline{3\arrayrulewidth}
\end{tabular}
}
\end{center}
\caption{The result of ablation experiment that excludes answer verification process from our framework.}
\label{tab:app-ablation}
\end{table}

\newpage
\section{Further Experimental Result}
\label{app:further-experimental-result}

\subsection{Domain Shift of Model in Image Captioning}
\begin{table}[h]
\begin{center}
\resizebox{\columnwidth}{!}{
\begin{tabular}{c|ccccc}
\Xhline{3\arrayrulewidth}
             & \multicolumn{5}{c}{\textbf{Trained on Cartoon Drawing}}                                                                                                                 \\
\textbf{ViT}                            & BLEU                         & ROUGE                        & METEOR                       & BERTS.                        & BARTS.                         \\ \hline\hline

\rowcolor{Gray}
\textit{Real}    & 21.64 & 35.07 & 21.82 & 0.5916 & -4.6258 \\ 
\textit{Cartoon} & 42.53 & 41.86 & 23.38 & 0.6721 & -4.8267 \\ 
\rowcolor{Gray}
\textit{Pencil}  & 31.50 & 35.63 & 18.79 & 0.6267 & -4.6112 \\ 
\rowcolor{Gray}
\textit{Oil}     & 30.66 & 33.39 & 17.32 & 0.6270 & -4.7253 \\ \Xhline{2\arrayrulewidth}
\textbf{CLIP}                           & BLEU                         & ROUGE                        & METEOR                       & BERTS.                        & BARTS.                         \\ \hline\hline
\rowcolor{Gray} \textit{Real}    & 20.20 & 32.90 & 19.23 & 0.5858 & -4.6253 \\ 
\textit{Cartoon} & 38.66 & 39.99 & 21.72 & 0.6595 & -4.8271 \\ 
\rowcolor{Gray} \textit{Pencil}  & 24.04 & 30.14 & 15.93 & 0.6036 & -4.6126 \\ 
\rowcolor{Gray} \textit{Oil}     & 27.69 & 30.97 & 15.77 & 0.6105 & -4.7255 \\ \Xhline{2\arrayrulewidth}
\textbf{BLIP}                           & BLEU                         & ROUGE                        & METEOR                       & BERTS.                        & BARTS.                         \\ \hline\hline
\rowcolor{Gray} \textit{Real}    & 21.21 & 32.60 & 22.96 & 0.5866 & -4.6698 \\ 
\textit{Cartoon} & 41.75 & 40.26 & 25.28 & 0.6822 & -4.8737 \\ 
\rowcolor{Gray} \textit{Pencil}  & 33.73 & 34.49 & 20.84 & 0.6313 & -4.6439 \\ 
\rowcolor{Gray} \textit{Oil}     & 34.83 & 34.71 & 18.92 & 0.6380 & -4.7294 \\ 
\Xhline{3\arrayrulewidth}
             & \multicolumn{5}{c}{\textbf{Trained on Pencil Drawing}}                                                                                                                 \\
\textbf{ViT}                            & BLEU                         & ROUGE                        & METEOR                       & BERTS.                        & BARTS.                         \\ \hline\hline
\rowcolor{Gray} \textit{Real}    & 21.61 & 34.18 & 22.31 & 0.5933 & -4.6253 \\ 
\rowcolor{Gray} \textit{Cartoon} & 35.50 & 38.05 & 20.66 & 0.6403 & -4.8264 \\ 
\textit{Pencil}  & 42.87 & 41.52 & 23.18 & 0.6481 & -4.6106 \\ 
\rowcolor{Gray} \textit{Oil}     & 33.92 & 34.56 & 18.38 & 0.6475 & -4.7253 \\ \Xhline{2\arrayrulewidth}
\textbf{CLIP}                           & BLEU                         & ROUGE                        & METEOR                       & BERTS.                        & BARTS.                         \\ \hline\hline
\rowcolor{Gray} \textit{Real}    & 20.73 & 32.72 & 18.59 & 0.5711 & -4.6253 \\ 
\rowcolor{Gray} \textit{Cartoon} & 30.28 & 34.02 & 17.64 & 0.6104 & -4.8264 \\ 
\textit{Pencil}  & 39.88 & 39.42 & 21.37 & 0.6298 & -4.6103 \\ 
\rowcolor{Gray} \textit{Oil}     & 30.67 & 32.22 & 16.52 & 0.6261 & -4.7253 \\ \Xhline{2\arrayrulewidth}
\textbf{BLIP}                           & BLEU                         & ROUGE                        & METEOR                       & BERTS.                        & BARTS.                         \\ \hline\hline
\rowcolor{Gray} \textit{Real}    & 19.18 & 29.52 & 22.64 & 0.5735 & -4.6752 \\ 
\rowcolor{Gray} \textit{Cartoon} & 34.30 & 34.04 & 21.47 & 0.6479 & -4.8780 \\ 
\textit{Pencil}  & 42.14 & 38.74 & 23.93 & 0.6537 & -4.6415 \\ 
\rowcolor{Gray} \textit{Oil}     & 33.97 & 33.41 & 19.17 & 0.6406 & -4.7284 \\ 
\Xhline{3\arrayrulewidth}
             & \multicolumn{5}{c}{\textbf{Trained on Oil Painting}}                                                                                                                 \\
\textbf{ViT}                            & BLEU                         & ROUGE                        & METEOR                       & BERTS.                        & BARTS.                         \\ \hline\hline
\rowcolor{Gray} \textit{Real}    & 19.20 & 29.74 & 21.41 & 0.5684 & -4.6254 \\ 
\rowcolor{Gray} \textit{Cartoon} & 33.76 & 35.58 & 21.37 & 0.6350 & -4.8274 \\ 
\rowcolor{Gray} \textit{Pencil}  & 34.34 & 34.85 & 20.60 & 0.6361 & -4.6111 \\ 
\textit{Oil}     & 46.97 & 42.39 & 23.75 & 0.6759 & -4.7253 \\ \Xhline{2\arrayrulewidth}
\textbf{CLIP}                           & BLEU                         & ROUGE                        & METEOR                       & BERTS.                        & BARTS.                         \\ \hline\hline
\rowcolor{Gray} \textit{Real}    & 19.33 & 29.83 & 20.30 & 0.5705 & -4.6251 \\ 
\rowcolor{Gray} \textit{Cartoon} & 32.13 & 34.14 & 19.56 & 0.6237 & -4.8262 \\
\rowcolor{Gray} \textit{Pencil}  & 31.34 & 32.51 & 18.35 & 0.6268 & -4.6103 \\
\textit{Oil}     & 46.11 & 42.09 & 23.05 & 0.6693 & -4.7253 \\ \Xhline{2\arrayrulewidth}
\textbf{BLIP}                           & BLEU                         & ROUGE                        & METEOR                       & BERTS.                        & BARTS.                         \\ \hline\hline
\rowcolor{Gray} \textit{Real}    & 21.12 & 30.38 & 22.26 & 0.5818 & -4.6253 \\ 
\rowcolor{Gray} \textit{Cartoon} & 34.41 & 36.09 & 21.20 & 0.6335 & -4.8264 \\ 
\rowcolor{Gray} \textit{Pencil}  & 35.30 & 35.32 & 20.55 & 0.6373 & -4.6105 \\ 
\textit{Oil}     & 46.67 & 41.18 & 25.01 & 0.6833 & -4.7306 \\ 
\Xhline{3\arrayrulewidth}
\end{tabular}
}
\end{center}
\caption{Supplementary experimental result demonstrating domain shift on image captioning task.}
\label{tab:shift-app-cap}
\end{table}

\onecolumn
\newpage
\subsection{Effectiveness of Domain Generalization in Image Captioning Task}
\begin{table*}[h]
\begin{center}
\resizebox{0.95\textwidth}{!}{
\begin{tabular}{c|ccccc|cccccc}
\Xhline{3\arrayrulewidth}
                                                      & \multicolumn{5}{c|}{\textbf{Trained on R+C}}                                                                                                                 & \multicolumn{6}{c}{\textbf{Trained on R+P}}                                                                                                                                                  \\
ViT                                                                     & BLEU  & ROUGE  & METEOR & BERTS.  & BARTS.  &          & BLEU  & ROUGE  & METEOR & BERTS.  & BARTS.  \\ \hline\hline
\textit{Real}                                             & 44.82 & 51.13 & 28.46 & 0.6838 & -4.6268 & \textit{Real}    & 44.74 & 51,14 & 28.30 & 0.6828 & -4.6261 \\
\textit{Cartoon}                                          & 41.95 & 41.64 & 23.35 & 0.6701 & -4.8279 & \cellcolor[gray]{0.9}\textit{Cartoon}  &\cellcolor[gray]{0.9} 36.19  &\cellcolor[gray]{0.9} 38.13  &\cellcolor[gray]{0.9} 20.56  &\cellcolor[gray]{0.9} 0.6410  &\cellcolor[gray]{0.9} -4.8279 \\
\cellcolor[gray]{0.9}\textit{Pencil}                                           & \cellcolor[gray]{0.9}31.48 & \cellcolor[gray]{0.9}36.07 & \cellcolor[gray]{0.9}18.76 & \cellcolor[gray]{0.9}0.6275 \cellcolor[gray]{0.9}& \cellcolor[gray]{0.9}-4.6130 &        \textit{Pencil}  & 43.01 & 41.91 & 23.28 & 0.6519 & -4.6116 \\
\cellcolor[gray]{0.9}\textit{Oil}                                              & \cellcolor[gray]{0.9}30.09 & \cellcolor[gray]{0.9}32.99 & \cellcolor[gray]{0.9}17.23 & \cellcolor[gray]{0.9}0.6281 & \cellcolor[gray]{0.9}-4.7254 & \cellcolor[gray]{0.9}\textit{Oil}      &\cellcolor[gray]{0.9} 31.08  &\cellcolor[gray]{0.9} 33.22  &\cellcolor[gray]{0.9} 17.68  &\cellcolor[gray]{0.9} 0.6287  &\cellcolor[gray]{0.9} -4.7277 \\ \Xhline{2\arrayrulewidth}
Frozen CLIP                                                             & BLEU  & ROUGE  & METEOR & BERTS.  & BARTS.  &          & BLEU  & ROUGE  & METEOR & BERTS.  & BARTS.  \\ \hline\hline
\textit{Real}                                             & 50.10 & 54.53 & 30.61 & 0.6977 & -4.6260 & \textit{Real}    & 49.20 & 54.48 & 30.54 & 0.6969 & -4.6253 \\
\textit{Cartoon}                                          & 42.06 & 41.87 & 23.35 & 0.6718 & -4.8264 & \cellcolor[gray]{0.9}\textit{Cartoon}  &\cellcolor[gray]{0.9} 35.69  &\cellcolor[gray]{0.9} 37.50  &\cellcolor[gray]{0.9} 20.61  &\cellcolor[gray]{0.9} 0.6353  &\cellcolor[gray]{0.9} -4.8264 \\
\cellcolor[gray]{0.9}\textit{Pencil}                                           & \cellcolor[gray]{0.9}30.64 & \cellcolor[gray]{0.9}35.48 & \cellcolor[gray]{0.9}18.75 & \cellcolor[gray]{0.9}0.6288 & \cellcolor[gray]{0.9}-4.6116 & \textit{Pencil}  & 43.51 & 41.17 & 23.02 & 0.6510 & -4.6105 \\
\cellcolor[gray]{0.9}\textit{Oil}                                              & \cellcolor[gray]{0.9}26.81 & \cellcolor[gray]{0.9}31.54 & \cellcolor[gray]{0.9}16.40 & \cellcolor[gray]{0.9}0.6170 & \cellcolor[gray]{0.9}-4.7253 & \cellcolor[gray]{0.9}\textit{Oil}      &\cellcolor[gray]{0.9} 29.12  &\cellcolor[gray]{0.9} 32.14  &\cellcolor[gray]{0.9} 17.66  &\cellcolor[gray]{0.9} 0.6222  &\cellcolor[gray]{0.9} -4.7253 \\ \Xhline{2\arrayrulewidth}
\begin{tabular}[c]{@{}c@{}}ViT\\ w/ \cite{ren2023crossing}\end{tabular} & BLEU  & ROUGE  & METEOR & BERTS.  & BARTS.  &          & BLEU  & ROUGE  & METEOR & BERTS.  & BARTS.  \\ \hline\hline
\textit{Real}                                             & 39.97 & 48.21 & 27.84 & 0.6769 & -4.6283 & \textit{Real}    & 46.96 & 35.16 & 29.50 & 0.6905 & -4.6284 \\
\textit{Cartoon}                                          & 42.38 & 42.04 & 23.04 & 0.6649 & -4.8264 & \cellcolor[gray]{0.9}\textit{Cartoon}  &\cellcolor[gray]{0.9} 36.65  &\cellcolor[gray]{0.9} 38.56  &\cellcolor[gray]{0.9} 21.61  &\cellcolor[gray]{0.9} 0.6406  &\cellcolor[gray]{0.9} -4.8236 \\
\cellcolor[gray]{0.9}\textit{Pencil}                                           & \cellcolor[gray]{0.9}31.72 & \cellcolor[gray]{0.9}36.38 & \cellcolor[gray]{0.9}19.17 & \cellcolor[gray]{0.9}0.6318 & \cellcolor[gray]{0.9}-4.6103 & \textit{Pencil}  & 42.05 & 40.79 & 23.23 & 0.6517 & -4.6124 \\
\cellcolor[gray]{0.9}\textit{Oil}                                              & \cellcolor[gray]{0.9}30.05 & \cellcolor[gray]{0.9}32.73 & \cellcolor[gray]{0.9}17.52 & \cellcolor[gray]{0.9}0.6306 & \cellcolor[gray]{0.9}-4.7235 & \cellcolor[gray]{0.9}\textit{Oil}      &\cellcolor[gray]{0.9} 31.59  &\cellcolor[gray]{0.9} 34.17  &\cellcolor[gray]{0.9} 17.70  &\cellcolor[gray]{0.9} 0.6414  &\cellcolor[gray]{0.9} -4.7235 \\ \Xhline{3\arrayrulewidth}
                                                      & \multicolumn{5}{c|}{\textbf{Trained on R+O}}                                                                                                                 & \multicolumn{6}{c}{\textbf{Trained on C+P}}                                                                                                                                                  \\
ViT                                                                     & BLEU  & ROUGE  & METEOR & BERTS.  & BARTS.  &          & BLEU  & ROUGE  & METEOR & BERTS.  & BARTS.  \\ \hline\hline
\textit{Real}                                             & 45.49 & 52.24 & 28.98 & 0.6947 & -4.6274 & \cellcolor[gray]{0.9}\textit{Real}     &\cellcolor[gray]{0.9} 20.65  &\cellcolor[gray]{0.9} 33.31  &\cellcolor[gray]{0.9} 22.14  &\cellcolor[gray]{0.9} 0.5762  &\cellcolor[gray]{0.9} -4.6293 \\
\cellcolor[gray]{0.9}\textit{Cartoon}                                           &\cellcolor[gray]{0.9} 32.04  &\cellcolor[gray]{0.9} 34.64  &\cellcolor[gray]{0.9} 19.10  &\cellcolor[gray]{0.9} 0.6265  &\cellcolor[gray]{0.9} -4.8269 & \textit{Cartoon} & 42.54 & 42.18 & 23.24 & 0.6616 & -4.8267 \\
\cellcolor[gray]{0.9}\textit{Pencil}                                            &\cellcolor[gray]{0.9} 32.73  &\cellcolor[gray]{0.9} 34.08  &\cellcolor[gray]{0.9} 18.66  &\cellcolor[gray]{0.9} 0.6272  &\cellcolor[gray]{0.9} -4.6134 & \textit{Pencil}  & 43.12 & 41.14 & 23.62 & 0.6469 & -4.6105 \\
\textit{Oil}                                              & 45.81 & 42.29 & 23.34 & 0.6749 & -4.7254 & \cellcolor[gray]{0.9}\textit{Oil}      &\cellcolor[gray]{0.9} 34.14  &\cellcolor[gray]{0.9} 34.79  &\cellcolor[gray]{0.9} 18.97  &\cellcolor[gray]{0.9} 0.6394  &\cellcolor[gray]{0.9} -4.7254 \\ \Xhline{2\arrayrulewidth}
Frozen CLIP                                                             & BLEU  & ROUGE  & METEOR & BERTS.  & BARTS.  &          & BLEU  & ROUGE  & METEOR & BERTS.  & BARTS.  \\ \hline\hline
\textit{Real}                                             & 48.64 & 53.52 & 30.03 & 0.6938 & -4.6292 & \cellcolor[gray]{0.9}\textit{Real}     &\cellcolor[gray]{0.9} 20.99  &\cellcolor[gray]{0.9} 34.66  &\cellcolor[gray]{0.9} 22.45  &\cellcolor[gray]{0.9} 0.5771  &\cellcolor[gray]{0.9} -4.6290 \\
\cellcolor[gray]{0.9}\textit{Cartoon} &                                         \cellcolor[gray]{0.9} 33.35 & \cellcolor[gray]{0.9} 34.42 &\cellcolor[gray]{0.9} 19.86 &\cellcolor[gray]{0.9} 0.6297 &\cellcolor[gray]{0.9} -4.8266 & \textit{Cartoon} & 42.82 & 42.20 & 23.64 & 0.6733 & -4.8264 \\
\cellcolor[gray]{0.9}\textit{Pencil}                                            &\cellcolor[gray]{0.9} 31.74  &\cellcolor[gray]{0.9} 34.53  &\cellcolor[gray]{0.9} 18.88  &\cellcolor[gray]{0.9} 0.6323  &\cellcolor[gray]{0.9} -4.6122 & \textit{Pencil}  & 43.09 & 41.64 & 23.72 & 0.6519 & -4.6103 \\
\textit{Oil}                                              & 45.68 & 42.04 & 23.05 & 0.6711 & -4.7353 & \cellcolor[gray]{0.9}\textit{Oil}      &\cellcolor[gray]{0.9} 34.50  &\cellcolor[gray]{0.9} 34.91  &\cellcolor[gray]{0.9} 19.12  &\cellcolor[gray]{0.9} 0.6353  &\cellcolor[gray]{0.9} -4.7255 \\ \Xhline{2\arrayrulewidth}
\begin{tabular}[c]{@{}c@{}}ViT\\ w/ \cite{ren2023crossing}\end{tabular} & BLEU  & ROUGE  & METEOR & BERTS.  & BARTS.  &          & BLEU  & ROUGE  & METEOR & BERTS.  & BARTS.  \\ \hline\hline
\textit{Real}                                             & 44.03 & 50.75 & 28.22 & 0.6811 & -4.6259 & \cellcolor[gray]{0.9}\textit{Real}     &\cellcolor[gray]{0.9} 23.89  &\cellcolor[gray]{0.9} 35.50  &\cellcolor[gray]{0.9} 22.89  &\cellcolor[gray]{0.9} 0.6353  &\cellcolor[gray]{0.9} -4.6253 \\
\cellcolor[gray]{0.9}\textit{Cartoon}                                           &\cellcolor[gray]{0.9} 34.42  &\cellcolor[gray]{0.9} 35.32  &\cellcolor[gray]{0.9} 20.75  &\cellcolor[gray]{0.9} 0.6366  &\cellcolor[gray]{0.9} -4.8264 & \textit{Cartoon} & 41.71 & 40.62 & 23.12 & 0.6445 & -4.8265 \\
\cellcolor[gray]{0.9}\textit{Pencil}                                            &\cellcolor[gray]{0.9} 34.99  &\cellcolor[gray]{0.9} 34.85  &\cellcolor[gray]{0.9} 19.92  &\cellcolor[gray]{0.9} 0.6324  &\cellcolor[gray]{0.9} -4.6112 & \textit{Pencil}  & 42.51 & 41.55 & 23.37 & 0.6483 & -4.6111 \\
\textit{Oil}                                              & 44.70 & 41.43 & 22.65 & 0.6721 & -4.7224 & \cellcolor[gray]{0.9}\textit{Oil}      &\cellcolor[gray]{0.9} 35.88  &\cellcolor[gray]{0.9} 35.42  &\cellcolor[gray]{0.9} 19.20  &\cellcolor[gray]{0.9} 0.6409  &\cellcolor[gray]{0.9} -4.7253 \\ \Xhline{3\arrayrulewidth}
                                                      & \multicolumn{5}{c|}{\textbf{Trained on C+O}}                                                                                                                 & \multicolumn{6}{c}{\textbf{Trained on R+P}}                                                                                                                                                  \\
ViT                                                                     & BLEU  & ROUGE  & METEOR & BERTS.  & BARTS.  &          & BLEU  & ROUGE  & METEOR & BERTS.  & BARTS.  \\ \hline\hline
\cellcolor[gray]{0.9}\textit{Real}                                              &\cellcolor[gray]{0.9} 20.66  &\cellcolor[gray]{0.9} 32.13  &\cellcolor[gray]{0.9} 22.10  &\cellcolor[gray]{0.9} 0.5763  &\cellcolor[gray]{0.9} -4.6265 & \textit{Real}    & 20.24 & 32.22 & 21.67 & 0.5778 & -4.6256 \\
\textit{Cartoon}                                          & 43.20 & 42.48 & 24.13 & 0.6781 & -4.8266 & \cellcolor[gray]{0.9}\textit{Cartoon}  &\cellcolor[gray]{0.9} 32.24  &\cellcolor[gray]{0.9} 34.71  &\cellcolor[gray]{0.9} 20.13  &\cellcolor[gray]{0.9} 0.6285  &\cellcolor[gray]{0.9} -4.8273 \\
\cellcolor[gray]{0.9}\textit{Pencil}                                            &\cellcolor[gray]{0.9} 35.17  &\cellcolor[gray]{0.9} 36.11  &\cellcolor[gray]{0.9} 20.14  &\cellcolor[gray]{0.9} 0.6316  &\cellcolor[gray]{0.9} -4.6109 & \textit{Pencil}  & 44.23 & 42.28 & 24.02 & 0.6547 & -4.6106 \\
\textit{Oil}                                              & 46.74 & 42.35 & 23.69 & 0.6789 & -4.7264 & \cellcolor[gray]{0.9}\textit{Oil}      &\cellcolor[gray]{0.9} 47.17  &\cellcolor[gray]{0.9} 43.04  &\cellcolor[gray]{0.9} 23.81  &\cellcolor[gray]{0.9} 0.6761  &\cellcolor[gray]{0.9} -4.7253 \\ \Xhline{2\arrayrulewidth}
Frozen CLIP                                                             & BLEU  & ROUGE  & METEOR & BERTS.  & BARTS.  &          & BLEU  & ROUGE  & METEOR & BERTS.  & BARTS.  \\ \hline\hline
\cellcolor[gray]{0.9}\textit{Real}                                              &\cellcolor[gray]{0.9} 20.62  &\cellcolor[gray]{0.9} 32.51  &\cellcolor[gray]{0.9} 22.62  &\cellcolor[gray]{0.9} 0.5815  &\cellcolor[gray]{0.9} -4.6257 & \textit{Real}    & 20.31 & 31.95 & 21.08 & 0.5782 & -4.6267 \\
\textit{Cartoon}                                          & 43.69 & 42.85 & 23.83 & 0.6816 & -4.8264 & \cellcolor[gray]{0.9}\textit{Cartoon}  &\cellcolor[gray]{0.9} 34.49  &\cellcolor[gray]{0.9} 37.22  &\cellcolor[gray]{0.9} 21.08  &\cellcolor[gray]{0.9} 0.6380  &\cellcolor[gray]{0.9} -4.8268 \\
\cellcolor[gray]{0.9} \textit{Pencil}                                            &\cellcolor[gray]{0.9} 35.60  &\cellcolor[gray]{0.9} 36.19  &\cellcolor[gray]{0.9} 20.49  &\cellcolor[gray]{0.9} 0.6345  &\cellcolor[gray]{0.9} -4.6159 & \textit{Pencil}  & 44.42 & 42.02 & 23.95 & 0.6549 & -4.6128 \\
\textit{Oil}                                              & 46.06 & 42.08 & 24.22 & 0.6801 & -4.7245 & \cellcolor[gray]{0.9}\textit{Oil}      &\cellcolor[gray]{0.9} 46.94  &\cellcolor[gray]{0.9} 43.37  &\cellcolor[gray]{0.9} 24.13  &\cellcolor[gray]{0.9} 0.6788  &\cellcolor[gray]{0.9} -4.7256 \\ \Xhline{2\arrayrulewidth}
\begin{tabular}[c]{@{}c@{}}ViT\\ w/ \cite{ren2023crossing}\end{tabular} & BLEU  & ROUGE  & METEOR & BERTS.  & BARTS.  &          & BLEU  & ROUGE  & METEOR & BERTS.  & BARTS.  \\ \hline\hline
\cellcolor[gray]{0.9}\textit{Real}                                              &\cellcolor[gray]{0.9} 22.57  &\cellcolor[gray]{0.9} 32.86  &\cellcolor[gray]{0.9} 22.76  &\cellcolor[gray]{0.9} 0.5829  &\cellcolor[gray]{0.9} -4.6253 & \textit{Real}    & 21.77 & 32.98 & 22.40 & 0.5839 & -4.6253 \\
\textit{Cartoon}                                          & 42.76 & 42.04 & 23.81 & 0.6751 & -4.8268 & \cellcolor[gray]{0.9}\textit{Cartoon}  &\cellcolor[gray]{0.9} 36.58  &\cellcolor[gray]{0.9} 38.23  &\cellcolor[gray]{0.9} 21.93  &\cellcolor[gray]{0.9} 0.6402  &\cellcolor[gray]{0.9} -4.8266 \\
\cellcolor[gray]{0.9}\textit{Pencil}                                            &\cellcolor[gray]{0.9} 36.83  &\cellcolor[gray]{0.9} 36.45  &\cellcolor[gray]{0.9} 20.75  &\cellcolor[gray]{0.9} 0.6429  &\cellcolor[gray]{0.9} -4.6103 & \textit{Pencil}  & 43.04 & 41.32 & 23.39 & 0.6498 & -4.6133 \\
\textit{Oil}                                              & 46.88 & 42.77 & 23.58 & 0.6766 & -4.7253 & \cellcolor[gray]{0.9}\textit{Oil}      &\cellcolor[gray]{0.9} 47.01  &\cellcolor[gray]{0.9} 42.80  &\cellcolor[gray]{0.9} 23.67  &\cellcolor[gray]{0.9} 0.6737  &\cellcolor[gray]{0.9} -4.7261 \\ \Xhline{3\arrayrulewidth}
\end{tabular}
}
\end{center}
\caption{Supplementary experimental result demonstrating the effectiveness of domain generalization technique on image captioning task. This table presents the result of the model trained with two source domains, instead of that of Table~\ref{tab:gen-cap} that leveraged three source domains.}
\label{tab:gen-app-cap}
\end{table*}

\newpage
\subsection{Experimental Results in Visual Question Answering Task}
\begin{table*}[h]
\begin{center}
\resizebox{\textwidth}{!}{
\begin{tabular}{c|cccc|cccc|cccc}
\Xhline{3\arrayrulewidth}
                                                                        & \multicolumn{4}{c|}{\textbf{Trained on Cartoon Drawing}}                                                                      & \multicolumn{4}{c|}{\textbf{Trained on Pencil Drawing}}                                                                       & \multicolumn{4}{c}{\textbf{Trained on Oil Painting}}                                                                         \\
\multirow{-2}{*}{\textbf{VQA}}                                          & \cellcolor[gray]{0.9}\textit{Real}  & \textit{Cartoon} & \cellcolor[gray]{0.9}\textit{Pencil} & \cellcolor[gray]{0.9}\textit{Oil}   & \cellcolor[gray]{0.9}\textit{Real}  & \cellcolor[gray]{0.9}\textit{Cartoon} & \textit{Pencil} & \cellcolor[gray]{0.9}\textit{Oil}   & \cellcolor[gray]{0.9}\textit{Real}  & \cellcolor[gray]{0.9}\textit{Cartoon} & \cellcolor[gray]{0.9}\textit{Pencil} & \textit{Oil}   \\ \hline\hline
ViT                                                                      &\cellcolor[gray]{0.9} 42.39 & 75.23    &\cellcolor[gray]{0.9} 67.88   &\cellcolor[gray]{0.9} 68.04  &\cellcolor[gray]{0.9} 41.79  &\cellcolor[gray]{0.9} 68.42   & 75.29   &\cellcolor[gray]{0.9} 65.29  &\cellcolor[gray]{0.9} 43.82  &\cellcolor[gray]{0.9} 61.56    &\cellcolor[gray]{0.9} 64.70  & 77.35 \\ 
CLIP                                                                     &\cellcolor[gray]{0.9} 44.72 & 76.47    &\cellcolor[gray]{0.9} 69.21   &\cellcolor[gray]{0.9} 67.64  &\cellcolor[gray]{0.9} 43.23  &\cellcolor[gray]{0.9} 68.19   & 75.88   &\cellcolor[gray]{0.9} 66.17  &\cellcolor[gray]{0.9} 44.41  &\cellcolor[gray]{0.9} 62.33    &\cellcolor[gray]{0.9} 65.84  & 78.82 \\ 
BLIP                                                                     &\cellcolor[gray]{0.9} 45.16 & 78.52    &\cellcolor[gray]{0.9} 68.92   &\cellcolor[gray]{0.9} 69.48  &\cellcolor[gray]{0.9} 43.58  &\cellcolor[gray]{0.9} 69.54   & 77.64   &\cellcolor[gray]{0.9} 67.53  &\cellcolor[gray]{0.9} 44.70  &\cellcolor[gray]{0.9} 63.41    &\cellcolor[gray]{0.9} 67.56  & 79.71 \\ 
\Xhline{2\arrayrulewidth}
                                                                        & \multicolumn{4}{c|}{\textbf{Trained on R+C}}                                                                                  & \multicolumn{4}{c|}{\textbf{Trained on R+P}}                                                                                  & \multicolumn{4}{c}{\textbf{Trained on R+O}}                                                                                  \\ 
\multirow{-2}{*}{\textbf{VQA}}                                          & \textit{Real}  & \textit{Cartoon}  &\cellcolor[gray]{0.9} \textit{Pencil}  &\cellcolor[gray]{0.9} \textit{Oil}   & \textit{Real}   &\cellcolor[gray]{0.9} \textit{Cartoon} & \textit{Pencil}  &\cellcolor[gray]{0.9} \textit{Oil}   & \textit{Real}   &\cellcolor[gray]{0.9} \textit{Cartoon}  &\cellcolor[gray]{0.9} \textit{Pencil} & \textit{Oil}   \\ \hline\hline
ViT                                                                     & 56.88 & 74.50    &\cellcolor[gray]{0.9} 73.88   &\cellcolor[gray]{0.9} 75.15 & 54.62  &\cellcolor[gray]{0.9} 72.79   & 76.82   &\cellcolor[gray]{0.9} 75.91 & 53.52  &\cellcolor[gray]{0.9} 72.53    &\cellcolor[gray]{0.9} 72.24  & 76.35 \\
Frozen CLIP                                                             & 54.68 & 74.76    &\cellcolor[gray]{0.9} 72.18   &\cellcolor[gray]{0.9} 75.06 & 55.68  &\cellcolor[gray]{0.9} 73.62   & 77.35   &\cellcolor[gray]{0.9} 75.88 & 55.10  &\cellcolor[gray]{0.9} 72.84    &\cellcolor[gray]{0.9} 70.21  & 76.93 \\ 
\begin{tabular}[c]{@{}c@{}}ViT\\ w/ \cite{ren2023crossing}\end{tabular} & 54.59 & 74.29    &\cellcolor[gray]{0.9} 74.53   &\cellcolor[gray]{0.9} 75.47 & 53.84  &\cellcolor[gray]{0.9} 74.24   & 76.24   &\cellcolor[gray]{0.9} 76.79 & 52.82  &\cellcolor[gray]{0.9} 73.79    &\cellcolor[gray]{0.9} 75.47  & 76.03 \\ 
\Xhline{2\arrayrulewidth}
                                                                        & \multicolumn{4}{c|}{\textbf{Trained on C+P}}                                                                                  & \multicolumn{4}{c|}{\textbf{Trained on C+O}}                                                                                  & \multicolumn{4}{c}{\textbf{Trained on P+O}}                                                                                  \\
\multirow{-2}{*}{\textbf{VQA}}                                           &\cellcolor[gray]{0.9} \textit{Real}  & \textit{Cartoon} & \textit{Pencil}  &\cellcolor[gray]{0.9} \textit{Oil}    &\cellcolor[gray]{0.9} \textit{Real}  & \textit{Cartoon}  &\cellcolor[gray]{0.9} \textit{Pencil} & \textit{Oil}    &\cellcolor[gray]{0.9} \textit{Real}   &\cellcolor[gray]{0.9} \textit{Cartoon} & \textit{Pencil} & \textit{Oil}   \\ \hline\hline
ViT                                                                      &\cellcolor[gray]{0.9} 44.53 & 74.44   & 75.76   &\cellcolor[gray]{0.9} 76.42  &\cellcolor[gray]{0.9} 45.12 & 74.82    &\cellcolor[gray]{0.9} 74.11  & 76.44  &\cellcolor[gray]{0.9} 44.85  &\cellcolor[gray]{0.9} 74.01   & 76.68  & 76.53 \\ 
Frozen CLIP                                                              &\cellcolor[gray]{0.9} 45.88 & 74.88   & 76.21   &\cellcolor[gray]{0.9} 76.56  &\cellcolor[gray]{0.9} 45.76 & 75.15    &\cellcolor[gray]{0.9} 74.32  & 76.38  &\cellcolor[gray]{0.9} 45.12  &\cellcolor[gray]{0.9} 74.15   & 77.05  & 77.03 \\ 
\begin{tabular}[c]{@{}c@{}}ViT\\ w/ \cite{ren2023crossing}\end{tabular}  &\cellcolor[gray]{0.9} 46.47 & 74.59   & 74.93   &\cellcolor[gray]{0.9} 76.94  &\cellcolor[gray]{0.9} 45.98 & 74.53    &\cellcolor[gray]{0.9} 75.29  & 77.06  &\cellcolor[gray]{0.9} 45.29  &\cellcolor[gray]{0.9} 74.88   & 76.53  & 76.47 \\ 
\Xhline{3\arrayrulewidth}
\end{tabular}
}
\end{center}
\caption{Supplementary experimental result demonstrating the domain shift and effectiveness of domain generalization technique on VQA task.}
\label{tab:app-vqa}
\end{table*}

\subsection{Experimental Results in Visual Entailment Task}
\begin{table*}[h]
\begin{center}
\resizebox{\textwidth}{!}{
\begin{tabular}{c|cccc|cccc|cccc}
\Xhline{3\arrayrulewidth}
                                                                        & \multicolumn{4}{c|}{\textbf{Trained on Cartoon Drawing}}                                                                      & \multicolumn{4}{c|}{\textbf{Trained on Pencil Drawing}}                                                                       & \multicolumn{4}{c}{\textbf{Trained on Oil Painting}}                                                                         \\
\multirow{-2}{*}{\textbf{VE}}                                            &\cellcolor[gray]{0.9} \textit{Real}  & \textit{Cartoon}  &\cellcolor[gray]{0.9} \textit{Pencil}  &\cellcolor[gray]{0.9} \textit{Oil}    &\cellcolor[gray]{0.9} \textit{Real}   &\cellcolor[gray]{0.9} \textit{Cartoon} & \textit{Pencil}  &\cellcolor[gray]{0.9} \textit{Oil}    &\cellcolor[gray]{0.9} \textit{Real}   &\cellcolor[gray]{0.9} \textit{Cartoon}  &\cellcolor[gray]{0.9} \textit{Pencil} & \textit{Oil}   \\ \hline\hline
ViT                                                                      &\cellcolor[gray]{0.9} 55.14 & 68.95    &\cellcolor[gray]{0.9} 64.01   &\cellcolor[gray]{0.9} 63.85  &\cellcolor[gray]{0.9} 55.32  &\cellcolor[gray]{0.9} 62.59   & 69.18   &\cellcolor[gray]{0.9} 65.16  &\cellcolor[gray]{0.9} 55.79  &\cellcolor[gray]{0.9} 63.43    &\cellcolor[gray]{0.9} 65.20  & 71.47 \\ 
CLIP                                                                     &\cellcolor[gray]{0.9} 54.56 & 69.81    &\cellcolor[gray]{0.9} 65.08   &\cellcolor[gray]{0.9} 64.49  &\cellcolor[gray]{0.9} 56.24  &\cellcolor[gray]{0.9} 63.84   & 69.70   &\cellcolor[gray]{0.9} 64.12  &\cellcolor[gray]{0.9} 55.23  &\cellcolor[gray]{0.9} 62.94    &\cellcolor[gray]{0.9} 64.93  & 71.87 \\
BLIP                                                                     &\cellcolor[gray]{0.9} 49.88 & 63.18    &\cellcolor[gray]{0.9} 58.29   &\cellcolor[gray]{0.9} 57.21  &\cellcolor[gray]{0.9} 48.15  &\cellcolor[gray]{0.9} 61.13   & 65.09   &\cellcolor[gray]{0.9} 61.04  &\cellcolor[gray]{0.9} 46.82  &\cellcolor[gray]{0.9} 60.21    &\cellcolor[gray]{0.9} 61.96  & 71.89 \\
\Xhline{2\arrayrulewidth}
                                                                        & \multicolumn{4}{c|}{\textbf{Trained on R+C}}                                                                                  & \multicolumn{4}{c|}{\textbf{Trained on R+P}}                                                                                  & \multicolumn{4}{c}{\textbf{Trained on R+O}}                                                                                  \\
\multirow{-2}{*}{\textbf{VE}}                                           & \textit{Real}  & \textit{Cartoon}  &\cellcolor[gray]{0.9} \textit{Pencil}  &\cellcolor[gray]{0.9} \textit{Oil}   & \textit{Real}   &\cellcolor[gray]{0.9} \textit{Cartoon} & \textit{Pencil}  &\cellcolor[gray]{0.9} \textit{Oil}   & \textit{Real}   &\cellcolor[gray]{0.9} \textit{Cartoon}  &\cellcolor[gray]{0.9} \textit{Pencil} & \textit{Oil}   \\ \hline\hline
ViT                                                                     & 72.83 & 69.89    &\cellcolor[gray]{0.9} 65.78   &\cellcolor[gray]{0.9} 65.52 & 69.70  &\cellcolor[gray]{0.9} 64.28   & 68.89   &\cellcolor[gray]{0.9} 65.38 & 72.02  &\cellcolor[gray]{0.9} 65.32    &\cellcolor[gray]{0.9} 64.71  & 67.96 \\
Frozen CLIP                                                             & 73.27 & 70.01    &\cellcolor[gray]{0.9} 65.24   &\cellcolor[gray]{0.9} 66.47 & 73.38  &\cellcolor[gray]{0.9} 66.83   & 69.10   &\cellcolor[gray]{0.9} 65.23 & 72.11  &\cellcolor[gray]{0.9} 64.21    &\cellcolor[gray]{0.9} 65.47  & 69.70 \\
\begin{tabular}[c]{@{}c@{}}ViT\\ w/ \cite{ren2023crossing}\end{tabular} & 72.51 & 69.71    &\cellcolor[gray]{0.9} 66.11   &\cellcolor[gray]{0.9} 67.72 & 72.60  &\cellcolor[gray]{0.9} 65.88   & 68.74   &\cellcolor[gray]{0.9} 67.47 & 72.17  &\cellcolor[gray]{0.9} 65.79    &\cellcolor[gray]{0.9} 66.51  & 68.77 \\
\Xhline{2\arrayrulewidth}
                                                                        & \multicolumn{4}{c|}{\textbf{Trained on C+P}}                                                                                  & \multicolumn{4}{c|}{\textbf{Trained on C+O}}                                                                                  & \multicolumn{4}{c}{\textbf{Trained on P+O}}                                                                                  \\
\multirow{-2}{*}{\textbf{VE}}                                            &\cellcolor[gray]{0.9} \textit{Real}  & \textit{Cartoon} & \textit{Pencil}  &\cellcolor[gray]{0.9} \textit{Oil}    &\cellcolor[gray]{0.9} \textit{Real}  & \textit{Cartoon}  &\cellcolor[gray]{0.9} \textit{Pencil} & \textit{Oil}    &\cellcolor[gray]{0.9} \textit{Real}   &\cellcolor[gray]{0.9} \textit{Cartoon} & \textit{Pencil} & \textit{Oil}   \\ \hline\hline
ViT                                                                      &\cellcolor[gray]{0.9} 55.28 & 70.55   & 68.66   &\cellcolor[gray]{0.9} 67.96  &\cellcolor[gray]{0.9} 55.97 & 70.63    &\cellcolor[gray]{0.9} 68.08  & 68.43  &\cellcolor[gray]{0.9} 55.06  &\cellcolor[gray]{0.9} 66.32   & 68.69  & 68.97 \\ 
Frozen CLIP                                                              &\cellcolor[gray]{0.9} 56.22 & 70.73   & 69.20   &\cellcolor[gray]{0.9} 66.89  &\cellcolor[gray]{0.9} 56.28 & 69.20    &\cellcolor[gray]{0.9} 67.89  & 70.62  &\cellcolor[gray]{0.9} 55.49  &\cellcolor[gray]{0.9} 67.55   & 69.12  & 70.37 \\ 
\begin{tabular}[c]{@{}c@{}}ViT\\ w/ \cite{ren2023crossing}\end{tabular}  &\cellcolor[gray]{0.9} 56.45 & 70.82   & 68.12   &\cellcolor[gray]{0.9} 68.57  &\cellcolor[gray]{0.9} 56.49 & 70.12    &\cellcolor[gray]{0.9} 68.31  & 69.44  &\cellcolor[gray]{0.9} 55.99  &\cellcolor[gray]{0.9} 68.04   & 68.61  & 68.85 \\
\Xhline{3\arrayrulewidth}
\end{tabular}
}
\end{center}
\caption{Supplementary experimental result demonstrating the domain shift and effectiveness of domain generalization technique on VE task.}
\label{tab:app-ve}
\end{table*}

\newpage
\section{Additional Examples of Annotated Data}
\label{app:additional-examples-annotation}

\subsection{Additional Examples on Image Prompt}
\label{app:additional-examples-annotation-prompt}
In this section, we present an original image $x_{\textit{ori}}$ with decomposed $p_{\textit{ori}}$ and its transformed versions $x_{\textit{sty}}$ and $p_{\textit{sty}}$ generated through the annotation process described in Section 3. The expressions for $p_{\textit{sty}}$ that contribute to the generation of the stylized image are boldfaced.

\subsubsection{Cartoon Drawing Style Images with Prompts}
\begin{table}[h!]
  \centering
  \resizebox{0.65\textwidth}{!}{
  \begin{tabular}{ m{.5\textwidth} | m{.5\textwidth} }
    \Xhline{3\arrayrulewidth}
    \multicolumn{1}{c|}{Original Image with $p_{\textit{ori}}$} & \multicolumn{1}{c}{Stylized Image with $p_{\textit{sty}}$} \\ \hline\hline
    
    \begin{center}
    \begin{minipage}{.2\textwidth}
    \includegraphics[width=\linewidth, height=\linewidth]{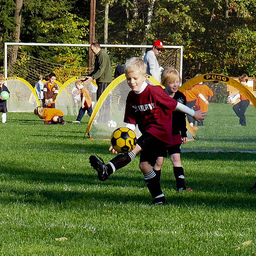}
    \end{minipage}
    \end{center}
    &
    \begin{center}
    \begin{minipage}{.2\textwidth}
    \includegraphics[width=\linewidth, height=\linewidth]{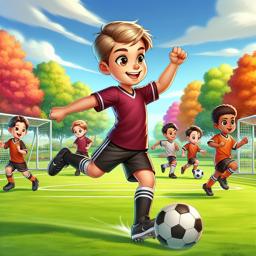}
    \end{minipage}
    \end{center}
    \\ 
    \begin{center}
    \begin{minipage}[t]{.5\textwidth}
    \small
    Create an image of a youth soccer game in a vibrant, well-lit outdoor setting. Focus on a young boy, wearing a maroon jersey, black shorts, and black socks with white stripes, as he kicks a yellow and black soccer ball. His hair is light and short, and he appears to be mid-action, showcasing a moment of dynamic movement. Surrounding him are other young children also playing soccer, dressed in various jerseys, primarily orange and white. The background features multiple small, arched soccer goals with a netting structure, hinting at a friendly, organized game environment. Lush green grass covers the field, and in the distance, trees with autumn foliage provide a scenic backdrop. The overall atmosphere should convey the excitement and energy of a youth soccer match on a sunny day.
    \end{minipage}
    \end{center}
    &
    \begin{center}
    \begin{minipage}[t]{.5\textwidth}
    \small
    Create a \textbf{cartoon-style} image of a youth soccer game in a vibrant, well-lit outdoor setting. Focus on a young boy, wearing a maroon jersey, black shorts, and black socks with white stripes, as he kicks a yellow and black soccer ball. His light, short hair and dynamic movement should be depicted with \textbf{exaggerated, playful features typical of cartoons}. Surrounding him are other young children also playing soccer, dressed in various jerseys, primarily orange and white. The background features multiple small, arched soccer goals with a netting structure, emphasizing a friendly, organized game environment. The field is illustrated with lush green grass, and in the distance, \textbf{colorful} autumn trees provide a scenic backdrop. The overall atmosphere should capture the excitement and energy of a youth soccer match on a sunny day, \textbf{with vibrant colors and whimsical details}.
    \end{minipage}
    \end{center}
    \\\\ \Xhline{2\arrayrulewidth}

    \begin{center}
    \begin{minipage}{.2\textwidth}
    \includegraphics[width=\linewidth, height=\linewidth]{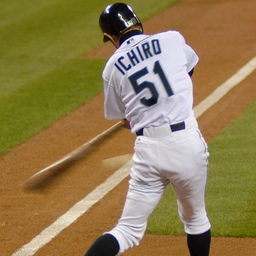}
    \end{minipage}
    \end{center}
    &
    \begin{center}
    \begin{minipage}{.2\textwidth}
    \includegraphics[width=\linewidth, height=\linewidth]{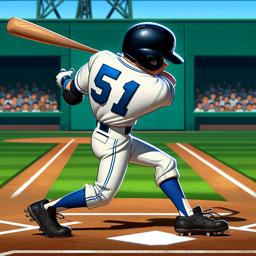}
    \end{minipage}
    \end{center}
    \\ 
    \begin{center}
    \begin{minipage}[t]{.5\textwidth}
    \small
    Create an image of a baseball player captured in mid-action swinging a bat. The player is wearing a white uniform with blue accents, notably with the number 51 and the name ``ICHIRO'' on the back. He is also wearing a black helmet, black socks, and black cleats. The scene takes place on a baseball field with a dirt basepath, green grass, and partially visible chalk lines. The player's stance and movement indicate a powerful swing, and his body is slightly bent forward with one leg stepping into the swing. The background should include the baseball field's elements subtly blurred to maintain the focus on the player. The overall atmosphere should convey the intensity and dynamism of the moment.
    \end{minipage}
    \end{center}
    &
    \begin{center}
    \begin{minipage}[t]{.5\textwidth}
    \small
    Create a \textbf{cartoon-style} image of a baseball player captured in mid-action swinging a bat. The player is depicted wearing a white uniform with blue accents, with the number 51 and the name ``ICHIRO'' on the back. He is also wearing a black helmet, black socks, and black cleats. The scene takes place on a cartoon baseball field with a dirt basepath, green grass, and \textbf{animated} chalk lines. The player's stance and movement indicate a powerful swing, and his body is slightly bent forward with one leg stepping into the swing. The background includes \textbf{stylized} elements of a baseball field, rendered with \textbf{exaggerated features and vibrant colors}, subtly blurred to maintain the focus on the player. The overall atmosphere should convey the intensity and dynamic action in a \textbf{whimsical, cartoonish manner}.
    \end{minipage} 
    \end{center}\\\\ \Xhline{2\arrayrulewidth}

    \begin{center}
    \begin{minipage}{.2\textwidth}
    \includegraphics[width=\linewidth, height=\linewidth]{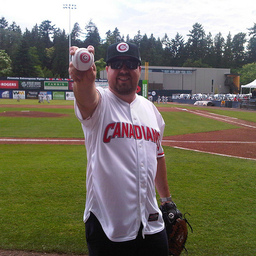}
    \end{minipage}
    \end{center}
    &
    \begin{center}
    \begin{minipage}{.2\textwidth}
    \includegraphics[width=\linewidth, height=\linewidth]{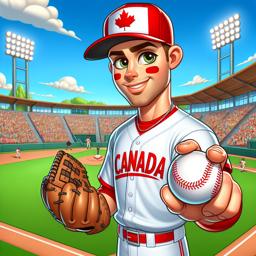}
    \end{minipage}
    \end{center}
    \\ 
    \begin{center}
    \begin{minipage}[t]{.5\textwidth}
    \small
    Create an image of a baseball player posing on a professional baseball field. The player is wearing a white baseball jersey with ``CANADA'' written across the chest in red letters, and a matching cap with a red maple leaf emblem. The player is pointing toward the camera with a baseball in hand, and their other hand is holding a glove. The background consists of a well-maintained baseball field, complete with bases, a pitcher's mound, and surrounding stadium seating filled with spectators. The weather is clear with scattered clouds, and lush green trees can be seen beyond the outfield. The atmosphere should be vibrant and playful, capturing the excitement of a baseball game day.
    \end{minipage}
    \end{center}
    &
    \begin{center}
    \begin{minipage}[t]{.5\textwidth}
    \small
    Create a \textbf{cartoon drawing style} image of a baseball player posing on a professional baseball field. The player is wearing a white baseball jersey with ``CANADA'' written across the chest in red letters, and a matching cap with a red maple leaf emblem. The player is pointing toward the camera with a baseball in hand, and their other hand is holding a glove. The background consists of a well-maintained baseball field, complete with bases, a pitcher's mound, and surrounding stadium seating filled with spectators. The weather is clear with scattered clouds, and lush green trees can be seen beyond the outfield. The atmosphere should be vibrant and playful, capturing the excitement of a baseball game day with \textbf{cartoonish, exaggerated features and vivid colors}.
    \end{minipage} 
    \end{center} \\\\ \Xhline{3\arrayrulewidth}
    
  \end{tabular}
  }
\end{table}

\newpage
\subsubsection{Pencil Drawing Style Images with Prompts}
\begin{table}[h!]
  \centering
  \resizebox{0.65\textwidth}{!}{
  \begin{tabular}{ m{.5\textwidth} | m{.5\textwidth} }
    \Xhline{3\arrayrulewidth}
    \multicolumn{1}{c|}{Original Image with $p_{\textit{ori}}$} & \multicolumn{1}{c}{Stylized Image with $p_{\textit{sty}}$} \\ \hline\hline
    
    \begin{center}
    \begin{minipage}{.2\textwidth}
    \includegraphics[width=\linewidth, height=\linewidth]{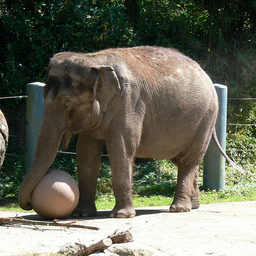}
    \end{minipage}
    \end{center}
    &
    \begin{center}
    \begin{minipage}{.2\textwidth}
    \includegraphics[width=\linewidth, height=\linewidth]{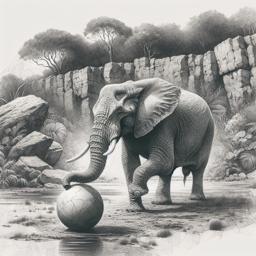}
    \end{minipage}
    \end{center}
    \\ 
    \begin{center}
    \begin{minipage}[t]{.5\textwidth}
    \small
    Create an image of an elephant in an outdoor setting, captured in a photorealistic style. The elephant is standing on a dirt ground with grass and vegetation in the background. It is interacting with a large ball, using its trunk to balance and play with the ball. The scene is set in a zoo or a wildlife sanctuary, with a sturdy fence in the background that includes metal posts and wires. Large rocks are present around the enclosure, enhancing the natural and rugged feel. The lighting suggests a sunny day, illuminating the elephant and casting shadows on the ground. The overall atmosphere should convey a sense of playful interaction and natural habitat for the elephant.
    \end{minipage}
    \end{center}
    &
    \begin{center}
    \begin{minipage}[t]{.5\textwidth}
    \small
    Create a \textbf{pencil drawing} of an elephant in an outdoor setting, captured with \textbf{fine details and shading}. The elephant is standing on a dirt ground with grass and vegetation in the background. It is interacting with a large ball, using its trunk to balance and play with it. The scene is set in a zoo or a wildlife sanctuary, depicted with a sturdy fence in the background that includes metal posts and wires. Large rocks are present around the enclosure, enhancing the natural and rugged feel. \textbf{The drawing should incorporate detailed cross-hatching and intricate line work to create depth and texture}. The overall atmosphere should convey a sense of playful interaction and natural habitat for the elephant.
    \end{minipage}
    \end{center}
    \\\\ \Xhline{2\arrayrulewidth}

    \begin{center}
    \begin{minipage}{.2\textwidth}
    \includegraphics[width=\linewidth, height=\linewidth]{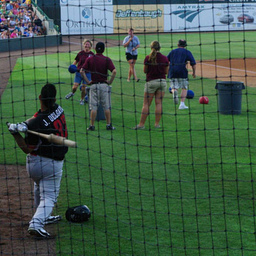}
    \end{minipage}
    \end{center}
    &
    \begin{center}
    \begin{minipage}{.2\textwidth}
    \includegraphics[width=\linewidth, height=\linewidth]{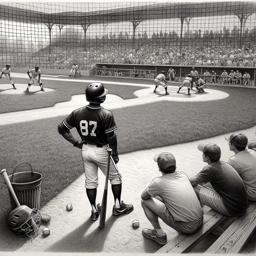}
    \end{minipage}
    \end{center}
    \\ 
    \begin{center}
    \begin{minipage}[t]{.5\textwidth}
    \small
    Create an image of a baseball game scene with a focus on a player standing near the edge of the field. The player is wearing a dark jersey, white pants, and a helmet, holding a bat under their arm. They are positioned on the dirt track looking toward the field, with a helmet resting on the ground nearby. The background features a baseball field with green grass, and several people are standing on the field, wearing casual attire consisting of T-shirts and shorts. There is a trash can and sports equipment scattered on the field. The stands in the background are filled with spectators. A netting separates the viewer from the action on the field, adding to the environment of a live baseball game. The atmosphere should capture the excitement and casual moments of a baseball game.
    \end{minipage}
    \end{center}
    &
    \begin{center}
    \begin{minipage}[t]{.5\textwidth}
    \small
    Create a \textbf{pencil drawing} of a baseball game scene with a focus on a player standing near the edge of the field. The player is wearing a dark jersey, white pants, and a helmet, holding a bat under their arm. They are positioned on the dirt track looking toward the field, with a helmet resting on the ground nearby. The background features a baseball field with green grass, and several people are standing on the field, wearing casual attire consisting of T-shirts and shorts. There is a trash can and sports equipment scattered on the field. The stands in the background are filled with spectators. A netting separates the viewer from the action on the field, adding to the environment of a live baseball game. \textbf{The entire scene should be rendered with detailed lines, cross-hatching, and shading to emphasize the pencil drawing style}, capturing the excitement and casual moments of a baseball game.
    \end{minipage} 
    \end{center}\\\\ \Xhline{2\arrayrulewidth}

    \begin{center}
    \begin{minipage}{.2\textwidth}
    \includegraphics[width=\linewidth, height=\linewidth]{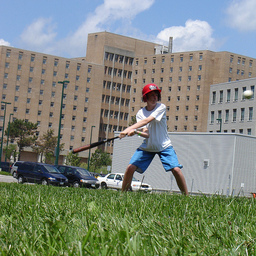}
    \end{minipage}
    \end{center}
    &
    \begin{center}
    \begin{minipage}{.2\textwidth}
    \includegraphics[width=\linewidth, height=\linewidth]{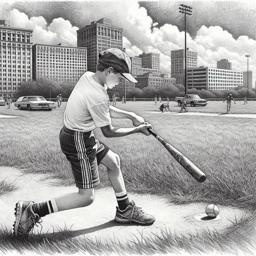}
    \end{minipage}
    \end{center}
    \\ 
    \begin{center}
    \begin{minipage}[t]{.5\textwidth}
    \small
    Create an image of a young boy playing baseball outdoors on a sunny day. The scene is set in a grassy field with tall buildings in the background. The boy is wearing a red baseball cap, a white t-shirt, and blue shorts. He is in mid-swing, holding a bat as if he's about to hit a baseball. The sky is clear with a few scattered clouds, and several cars are parked near the buildings in the background. The overall atmosphere should capture a warm, sunny day with the boy engrossed in his game amidst an urban setting. The image should reflect a casual, recreational vibe, emphasizing the contrast between the natural, grassy field and the tall, urban buildings.
    \end{minipage}
    \end{center}
    &
    \begin{center}
    \begin{minipage}[t]{.5\textwidth}
    \small
    Create a \textbf{pencil drawing} of a young boy playing baseball outdoors on a sunny day. The scene is set in a grassy field with tall buildings in the background. The boy is wearing a red baseball cap, a white t-shirt, and blue shorts. He is in mid-swing, holding a bat as if he's about to hit a baseball. The sky is clear with a few scattered clouds, and several cars are parked near the buildings in the background. The overall atmosphere should capture a warm, sunny day with the boy engrossed in his game amidst an urban setting. \textbf{The pencil drawing should include detailed line work, shading, and cross-hatching to give depth and texture}, capturing the contrast between the natural, grassy field and the tall, urban buildings.
    \end{minipage} 
    \end{center} \\\\ \Xhline{3\arrayrulewidth}
    
  \end{tabular}
  }
\end{table}

\newpage
\subsubsection{Oil Painting Style Images with Prompts}
\begin{table}[h!]
  \centering
  \resizebox{0.65\textwidth}{!}{
  \begin{tabular}{ m{.5\textwidth} | m{.5\textwidth} }
    \Xhline{3\arrayrulewidth}
    \multicolumn{1}{c|}{Original Image with $p_{\textit{ori}}$} & \multicolumn{1}{c}{Stylized Image with $p_{\textit{sty}}$} \\ \hline\hline
    
    \begin{center}
    \begin{minipage}{.2\textwidth}
    \includegraphics[width=\linewidth, height=\linewidth]{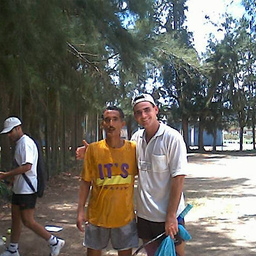}
    \end{minipage}
    \end{center}
    &
    \begin{center}
    \begin{minipage}{.2\textwidth}
    \includegraphics[width=\linewidth, height=\linewidth]{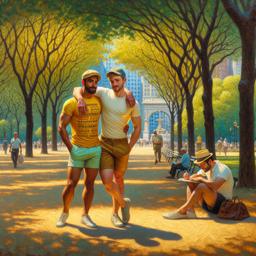}
    \end{minipage}
    \end{center}
    \\ 
    \begin{center}
    \begin{minipage}[t]{.5\textwidth}
    \small
    Create an image set in an outdoor park under the shade of tall, green trees. In the foreground, capture two men standing close to each other. The man on the left is wearing a yellow t-shirt with bold text across the chest and light-colored shorts. The man on the right is wearing a white shirt and shorts, along with a white cap, and has his arm around the other man's shoulder. Both are smiling or displaying a friendly demeanor. In the background, depict a third man wearing a white shirt and black shorts, also with a cap, who is walking away while looking down at something in his hands, such as a notebook or a phone. The ground is a mix of dirt and grass, and the sunlight filters through the tree branches, creating dappled shadows. Additionally, there should be some buildings or structures faintly visible further back, adding depth to the scene. The overall atmosphere should be casual and friendly, suggesting a leisurely day in the park.
    \end{minipage}
    \end{center}
    &
    \begin{center}
    \begin{minipage}[t]{.5\textwidth}
    \small
    Create an \textbf{oil painting} set in an outdoor park under the shade of tall, green trees. In the foreground, depict two men standing close to each other with warm, textured brushstrokes. The man on the left is wearing a yellow t-shirt with bold text across the chest and light-colored shorts, and the man on the right is wearing a white shirt, shorts, and a white cap, with his arm around the other man's shoulder. Both are portrayed with friendly expressions, \textbf{rendered with rich, vivid colors typical of oil paintings}. In the background, include a third man in a white shirt and black shorts, also wearing a cap, walking away while looking down at something in his hands, such as a notebook or a phone. The ground should be \textbf{painted} with a mix of earthy tones depicting dirt and grass, with sunlight filtering through the tree branches, creating dappled shadows. Additionally, some buildings or structures should be faintly visible further back, adding depth to the scene. The overall atmosphere should convey a casual and friendly leisurely day in the park, \textbf{with the warmth and depth characteristic of an oil painting}.
    \end{minipage}
    \end{center}
    \\\\ \Xhline{2\arrayrulewidth}

    \begin{center}
    \begin{minipage}{.2\textwidth}
    \includegraphics[width=\linewidth, height=\linewidth]{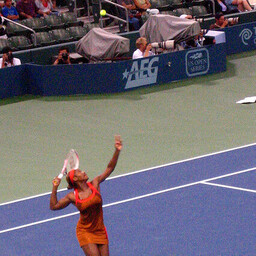}
    \end{minipage}
    \end{center}
    &
    \begin{center}
    \begin{minipage}{.2\textwidth}
    \includegraphics[width=\linewidth, height=\linewidth]{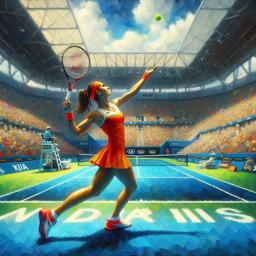}
    \end{minipage}
    \end{center}
    \\ 
    \begin{center}
    \begin{minipage}[t]{.5\textwidth}
    \small
    Create an image of a tennis player in the middle of executing a powerful serve during a match. The player is wearing a bright orange outfit and white shoes, with a red headband. The player is positioned on a blue and green tennis court, with one arm extended upward, holding the racquet ready to strike the ball. The stance and motion should convey intensity and athleticism. In the background, there is a stadium filled with spectators watching the match, with some sections covered by tarps. The scene should capture the dynamic energy and focus of a professional tennis match in a large, well-lit arena.
    \end{minipage}
    \end{center}
    &
    \begin{center}
    \begin{minipage}[t]{.5\textwidth}
    \small
    Create an \textbf{oil painting} of a tennis player in the middle of executing a powerful serve during a match. The player is wearing a bright orange outfit and white shoes, with a red headband, \textbf{all depicted with the textured brushstrokes and rich colors characteristic of oil painting.} The player is positioned on a \textbf{vibrant} blue and green tennis court, with one arm extended upward, holding the racquet ready to strike the ball. The stance and motion should convey intensity and athleticism, \textbf{captured with dynamic brushwork}. In the background, a stadium filled with spectators is \textbf{illustrated with a blend of detailed and impressionistic techniques}, showcasing their engagement and anticipation. Some sections of the stands are covered by tarps. The scene should evoke the dynamic energy and focus of a professional tennis match in a large, well-lit arena, \textbf{with an emphasis on the vivid, expressive style of an oil painting}.
    \end{minipage} 
    \end{center}\\\\ \Xhline{2\arrayrulewidth}

    \begin{center}
    \begin{minipage}{.2\textwidth}
    \includegraphics[width=\linewidth, height=\linewidth]{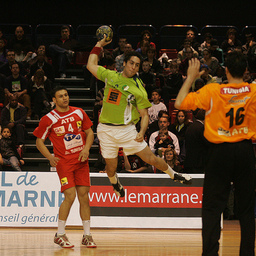}
    \end{minipage}
    \end{center}
    &
    \begin{center}
    \begin{minipage}{.2\textwidth}
    \includegraphics[width=\linewidth, height=\linewidth]{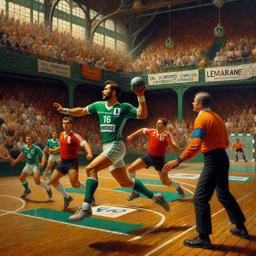}
    \end{minipage}
    \end{center}
    \\ 
    \begin{center}
    \begin{minipage}[t]{.5\textwidth}
    \small
    Create an image of a dynamic indoor handball match in progress. In the foreground, a player in a bright green jersey and white shorts is captured in mid-air as he attempts a powerful shot at the goal. He holds the ball in his right hand, showcasing his athleticism. To his left, two players dressed in red jerseys and white shorts are intensely focused on the play, one of them actively engaged in defense. In the right foreground, a referee in an orange shirt and black pants, with the number 16 on his back, is standing with his whistle ready to ensure fair play. The crowd in the background is seated in a dimly lit arena, watching the action with keen interest. Prominent banners and advertisements, including one with the text ``VAL de MARNE Conseil général'' and another for ``lemarrane.com,'' are displayed along the sides of the court, enhancing the realistic atmosphere of a professional handball game. The flooring is a polished wooden surface, capturing the energy and intensity of the match.
    \end{minipage}
    \end{center}
    &
    \begin{center}
    \begin{minipage}[t]{.5\textwidth}
    \small
    Create an \textbf{oil painting} of a dynamic indoor handball match in progress. In the foreground, a player in a bright green jersey and white shorts is depicted in mid-air, attempting a powerful shot at the goal with the ball in his right hand. The \textbf{painting} should capture his athleticism and motion with \textbf{expressive brushstrokes}. To his left, two players in red jerseys and white shorts are intensely focused on the play, one of them actively engaged in defense. On the right, a referee in an orange shirt and black pants, with the number 16 on his back, stands with his whistle ready to ensure fair play. The crowd in the background is seated in a dimly lit arena, watching the action with keen interest, \textbf{rendered with artistic details}. Prominent banners and advertisements, including one with the text ``VAL de MARNE Conseil général'' and another for ``lemarrane.com,'' are \textbf{painted} along the sides of the court, enhancing the realistic atmosphere of a professional handball game. The polished wooden flooring should be depicted with \textbf{rich textures}, capturing the energy and intensity of the match \textbf{through the depth and warmth typical of an oil painting}.
    \end{minipage} 
    \end{center} \\\\ \Xhline{3\arrayrulewidth}
    
  \end{tabular}
  }
\end{table}

\newpage
\subsection{Additional Examples on Image Captioning Task}
\begin{table}[h!]
  \centering
  \resizebox{0.95\textwidth}{!}{
  \begin{tabular}{ m{.5\textwidth} | m{.5\textwidth} }
    \Xhline{3\arrayrulewidth}
    \multicolumn{1}{c|}{Original Data} & \multicolumn{1}{c}{Annotated Data} \\ \hline\hline
    
    \begin{center}
    \begin{minipage}{.2\textwidth}
    \includegraphics[width=\linewidth, height=\linewidth]{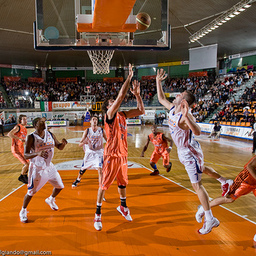}
    \end{minipage}
    \end{center}
    &
    \begin{center}
    \begin{minipage}{.2\textwidth}
    \includegraphics[width=\linewidth, height=\linewidth]{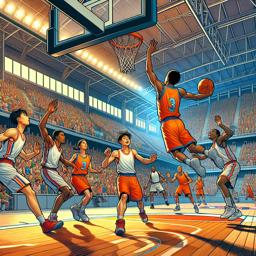}
    \end{minipage}
    \end{center}
    \\ 
    \begin{center}
    \begin{minipage}[t]{.5\textwidth}
    \tiny
      \begin{itemize}
        \item A group of basketball players on court during a game
        \item Basketball players in the process of making and defending a basket during a basketball game in an arena.
        \item A group of basketball players in the court as crowd looks
        \item Some men playing basketball with some fans watching
        \item A group of men playing basketball against each other.
      \end{itemize}
    \end{minipage}
    \end{center}
    &
    \begin{center}
    \begin{minipage}[t]{.5\textwidth}
    \tiny
      \begin{itemize}
        \item A lively cartoon scene of basketball players on the court during an intense game with a packed arena.
        \item Animated basketball players in mid-action, defending and attempting a shot in a vibrant, crowded indoor arena.
        \item Cartoon-style basketball players energetically competing on the court as a colorful crowd watches.
        \item Dynamic image of men playing basketball in an animated style, with enthusiastic fans cheering in the background.
        \item Animated depiction of a group of men engaged in a basketball game, surrounded by a lively audience in a large arena.
      \end{itemize}
    \end{minipage}
    \end{center}
    \\\\ \Xhline{2\arrayrulewidth}

    \begin{center}
    \begin{minipage}{.2\textwidth}
    \includegraphics[width=\linewidth, height=\linewidth]{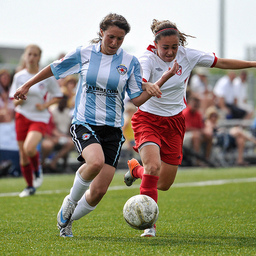}
    \end{minipage}
    \end{center}
    &
    \begin{center}
    \begin{minipage}{.2\textwidth}
    \includegraphics[width=\linewidth, height=\linewidth]{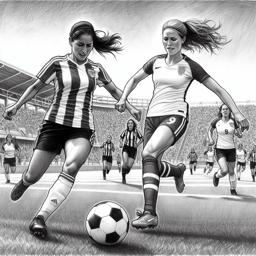}
    \end{minipage}
    \end{center}
    \\ 
    \begin{center}
    \begin{minipage}[t]{.5\textwidth}
    \tiny
      \begin{itemize}
        \item A group of girls on a field playing soccer.
        \item A group of women playing soccer on field with people watching.
        \item Two women chasing after a soccer ball on a field.
        \item Two girls on opposite teams competing for the soccer ball.
        \item Two teams playing soccer while people are watching.
      \end{itemize}
    \end{minipage}
    \end{center}
    &
    \begin{center}
    \begin{minipage}[t]{.5\textwidth}
    \tiny
      \begin{itemize}
        \item Two female soccer players in dynamic motion as they compete for the ball on a crowded field.
        \item An intense women's soccer match, skillfully illustrated in pencil, with spectators cheering in the background.
        \item Two determined athletes from opposing teams vying for control of the ball during a fierce soccer game.
        \item A high-energy soccer match with two women battling for possession, surrounded by an enthusiastic crowd.
        \item A competitive soccer scene, showing two women in action and an audience engrossed in the game, all rendered in intricate pencil detail.
      \end{itemize}
    \end{minipage} 
    \end{center}\\\\ \Xhline{2\arrayrulewidth}

    \begin{center}
    \begin{minipage}{.2\textwidth}
    \includegraphics[width=\linewidth, height=\linewidth]{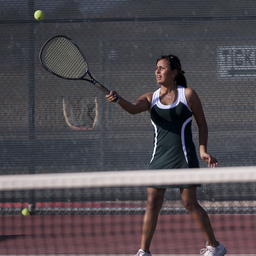}
    \end{minipage}
    \end{center}
    &
    \begin{center}
    \begin{minipage}{.2\textwidth}
    \includegraphics[width=\linewidth, height=\linewidth]{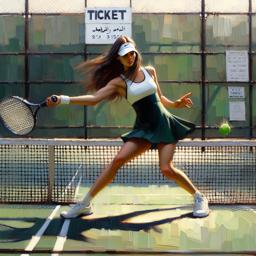}
    \end{minipage}
    \end{center}
    \\ 
    \begin{center}
    \begin{minipage}[t]{.5\textwidth}
    \tiny
      \begin{itemize}
        \item A woman swings her tennis racket at a tennis ball.
        \item A lady wearing white shoes and in a black outfit  is playing tennis.
        \item A woman extends her arm to hit a tennis ball.
        \item A beautiful young woman hitting a tennis ball with a racquet.
        \item A woman in a green tennis dress and white sneakers playing tennis on a court.
      \end{itemize}
    \end{minipage}
    \end{center}
    &
    \begin{center}
    \begin{minipage}[t]{.5\textwidth}
    \tiny
      \begin{itemize}
        \item A woman in a green and white tennis dress swings her racket at a tennis ball, captured in a vibrant oil painting style.
        \item An athlete, wearing white sneakers and a dark green outfit, is painted mid-action while playing tennis.
        \item A depiction of a woman extending her arm to strike a tennis ball with dynamic brushwork.
        \item A beautiful young woman hits a tennis ball with a racket in an oil-painted scene.
        \item On an outdoor court, a woman in a green tennis dress and white sneakers engages in a tennis match, rendered with lush, textured strokes.
      \end{itemize}
    \end{minipage} 
    \end{center} \\\\ \Xhline{3\arrayrulewidth}
    
  \end{tabular}
  }
\end{table}

\newpage
\subsection{Additional Examples on Visual Question Answering Task}
\label{app:additional-examples-annotation-vqa}
\begin{table}[h!]
  \centering
  \resizebox{0.85\textwidth}{!}{
  \begin{tabular}{ m{.25\textwidth} | m{.25\textwidth} | m{.25\textwidth} | m{.25\textwidth} }
    \Xhline{3\arrayrulewidth}
    \multicolumn{1}{c|}{Original Data} & \multicolumn{1}{c|}{Annotated Data} & \multicolumn{1}{c|}{Original Data} & \multicolumn{1}{c}{Annotated Data} \\ \hline\hline

    \begin{center}
    \begin{minipage}{.1\textwidth}
    \includegraphics[width=\linewidth, height=\linewidth]{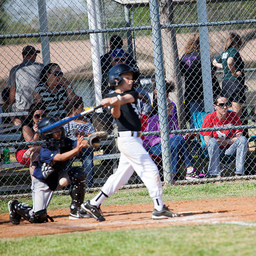}
    \end{minipage}
    \end{center}
    &
    \begin{center}
    \begin{minipage}{.1\textwidth}
    \includegraphics[width=\linewidth, height=\linewidth]{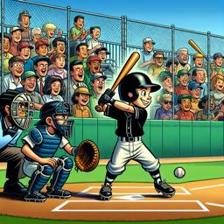}
    \end{minipage}
    \end{center}
    &
    \begin{center}
    \begin{minipage}{.1\textwidth}
    \includegraphics[width=\linewidth, height=\linewidth]{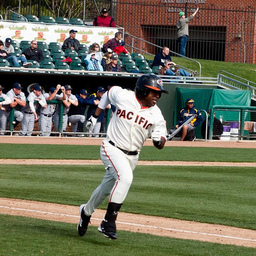}
    \end{minipage}
    \end{center}
    &
    \begin{center}
    \begin{minipage}{.1\textwidth}
    \includegraphics[width=\linewidth, height=\linewidth]{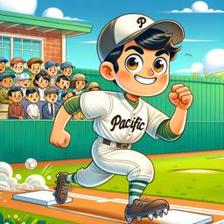}
    \end{minipage}
    \end{center}
    \\

    \begin{center}
    \begin{minipage}[t]{.25\textwidth}
    \small
      \begin{itemize}
        \item Question: Did he hit that ball?
        \item Answer: No
      \end{itemize}
    \end{minipage}
    \end{center}
    &
    \begin{center}
    \begin{minipage}[t]{.25\textwidth}
    \small
      \begin{itemize}
        \item Question: Did he strike the ball?
        \item Answer: No
      \end{itemize}
    \end{minipage}
    \end{center}
    &
    \begin{center}
    \begin{minipage}[t]{.25\textwidth}
    \small
      \begin{itemize}
        \item Question: Did a lot of people show up for the game?
        \item Answer: No
      \end{itemize}
    \end{minipage}
    \end{center}
    &
    \begin{center}
    \begin{minipage}[t]{.25\textwidth}
    \small
      \begin{itemize}
        \item Question: Was there a large crowd at the game?
        \item Answer: No
      \end{itemize}
    \end{minipage}
    \end{center}
    \\ \Xhline{2\arrayrulewidth}

    \begin{center}
    \begin{minipage}{.1\textwidth}
    \includegraphics[width=\linewidth, height=\linewidth]{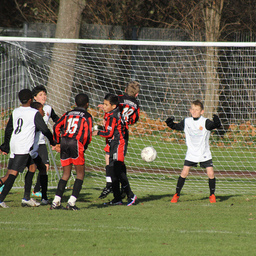}
    \end{minipage}
    \end{center}
    &
    \begin{center}
    \begin{minipage}{.1\textwidth}
    \includegraphics[width=\linewidth, height=\linewidth]{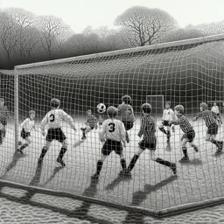}
    \end{minipage}
    \end{center}
    &
    \begin{center}
    \begin{minipage}{.1\textwidth}
    \includegraphics[width=\linewidth, height=\linewidth]{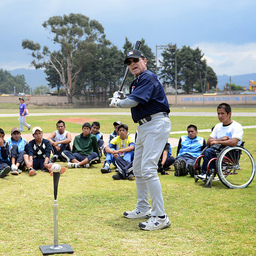}
    \end{minipage}
    \end{center}
    &
    \begin{center}
    \begin{minipage}{.1\textwidth}
    \includegraphics[width=\linewidth, height=\linewidth]{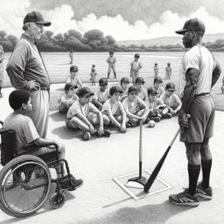}
    \end{minipage}
    \end{center}
    \\

    \begin{center}
    \begin{minipage}[t]{.25\textwidth}
    \small
      \begin{itemize}
        \item Question: Does the boy have his head stuck in the net?
        \item Answer: No
      \end{itemize}
    \end{minipage}
    \end{center}
    &
    \begin{center}
    \begin{minipage}[t]{.25\textwidth}
    \small
      \begin{itemize}
        \item Question: Is the boy's head caught in the net?
        \item Answer: No
      \end{itemize}
    \end{minipage}
    \end{center}
    &
    \begin{center}
    \begin{minipage}[t]{.25\textwidth}
    \small
      \begin{itemize}
        \item Question: Is there a disabled person?
        \item Answer: Yes
      \end{itemize}
    \end{minipage}
    \end{center}
    &
    \begin{center}
    \begin{minipage}[t]{.25\textwidth}
    \small
      \begin{itemize}
        \item Question: Is there a person with a disability?
        \item Answer: Yes
      \end{itemize}
    \end{minipage}
    \end{center}
    \\ \Xhline{2\arrayrulewidth}

    \begin{center}
    \begin{minipage}{.1\textwidth}
    \includegraphics[width=\linewidth, height=\linewidth]{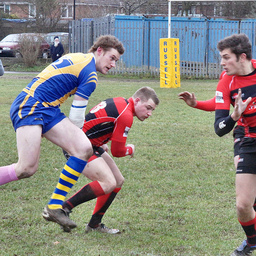}
    \end{minipage}
    \end{center}
    &
    \begin{center}
    \begin{minipage}{.1\textwidth}
    \includegraphics[width=\linewidth, height=\linewidth]{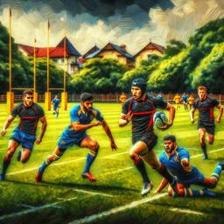}
    \end{minipage}
    \end{center}
    &
    \begin{center}
    \begin{minipage}{.1\textwidth}
    \includegraphics[width=\linewidth, height=\linewidth]{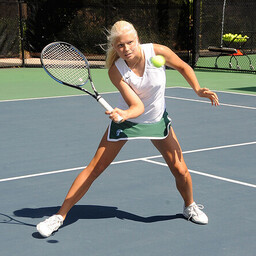}
    \end{minipage}
    \end{center}
    &
    \begin{center}
    \begin{minipage}{.1\textwidth}
    \includegraphics[width=\linewidth, height=\linewidth]{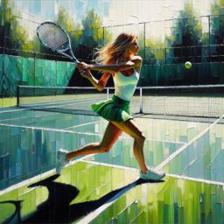}
    \end{minipage}
    \end{center}
    \\

    \begin{center}
    \begin{minipage}[t]{.25\textwidth}
    \small
      \begin{itemize}
        \item Question: Are the guys in blue wearing two different socks?
        \item Answer: Yes
      \end{itemize}
    \end{minipage}
    \end{center}
    &
    \begin{center}
    \begin{minipage}[t]{.25\textwidth}
    \small
      \begin{itemize}
        \item Question: Do the men in blue have mismatched socks?
        \item Answer: No
      \end{itemize}
    \end{minipage}
    \end{center}
    &
    \begin{center}
    \begin{minipage}[t]{.25\textwidth}
    \small
      \begin{itemize}
        \item Question: Is the girls right arm in an awkward position? 
        \item Answer: Yes
      \end{itemize}
    \end{minipage}
    \end{center}
    &
    \begin{center}
    \begin{minipage}[t]{.25\textwidth}
    \small
      \begin{itemize}
        \item Question: Is the girl's right arm positioned awkwardly? 
        \item Answer: No
      \end{itemize}
    \end{minipage}
    \end{center}
    \\ \Xhline{3\arrayrulewidth}
    
  \end{tabular}
  }
\end{table}

\subsection{Additional Examples on Visual Entailment Task}
\begin{table}[h!]
  \centering
  \resizebox{0.85\textwidth}{!}{
  \begin{tabular}{ m{.25\textwidth} | m{.25\textwidth} | m{.25\textwidth} | m{.25\textwidth} }
    \Xhline{3\arrayrulewidth}
    \multicolumn{1}{c|}{Original Data} & \multicolumn{1}{c|}{Annotated Data} & \multicolumn{1}{c|}{Original Data} & \multicolumn{1}{c}{Annotated Data} \\ \hline\hline

    \begin{center}
    \begin{minipage}{.1\textwidth}
    \includegraphics[width=\linewidth, height=\linewidth]{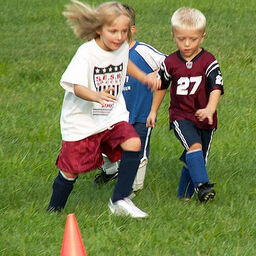}
    \end{minipage}
    \end{center}
    &
    \begin{center}
    \begin{minipage}{.1\textwidth}
    \includegraphics[width=\linewidth, height=\linewidth]{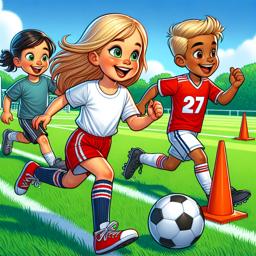}
    \end{minipage}
    \end{center}
    &
    \begin{center}
    \begin{minipage}{.1\textwidth}
    \includegraphics[width=\linewidth, height=\linewidth]{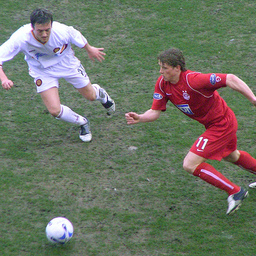}
    \end{minipage}
    \end{center}
    &
    \begin{center}
    \begin{minipage}{.1\textwidth}
    \includegraphics[width=\linewidth, height=\linewidth]{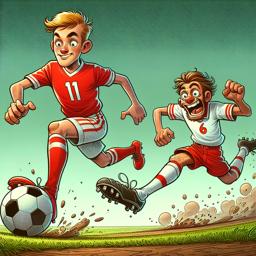}
    \end{minipage}
    \end{center}
    \\

    \begin{center}
    \begin{minipage}[t]{.25\textwidth}
    \small
      \begin{itemize}
        \item Hypothesis: Adults are playing frisbee 
        \item Label: Contradiction
      \end{itemize}
    \end{minipage}
    \end{center}
    &
    \begin{center}
    \begin{minipage}[t]{.25\textwidth}
    \small
      \begin{itemize}
        \item Hypothesis: Grown-ups are tossing a frisbee around.
        \item Label: Contradiction
      \end{itemize}
    \end{minipage}
    \end{center}
    &
    \begin{center}
    \begin{minipage}[t]{.25\textwidth}
    \small
      \begin{itemize}
        \item Hypothesis: Two sports players are sprinting towards the ball.
        \item Label: Entailment
      \end{itemize}
    \end{minipage}
    \end{center}
    &
    \begin{center}
    \begin{minipage}[t]{.25\textwidth}
    \small
      \begin{itemize}
        \item Hypothesis: Two athletes are racing toward the ball.
        \item Label: Entailment
      \end{itemize}
    \end{minipage}
    \end{center}
    \\ \Xhline{2\arrayrulewidth}

    \begin{center}
    \begin{minipage}{.1\textwidth}
    \includegraphics[width=\linewidth, height=\linewidth]{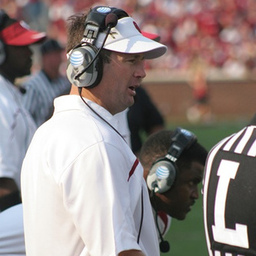}
    \end{minipage}
    \end{center}
    &
    \begin{center}
    \begin{minipage}{.1\textwidth}
    \includegraphics[width=\linewidth, height=\linewidth]{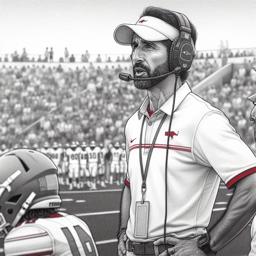}
    \end{minipage}
    \end{center}
    &
    \begin{center}
    \begin{minipage}{.1\textwidth}
    \includegraphics[width=\linewidth, height=\linewidth]{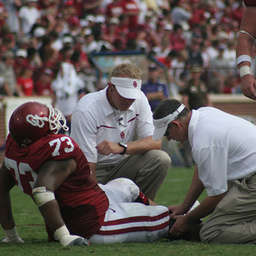}
    \end{minipage}
    \end{center}
    &
    \begin{center}
    \begin{minipage}{.1\textwidth}
    \includegraphics[width=\linewidth, height=\linewidth]{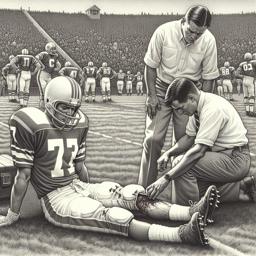}
    \end{minipage}
    \end{center}
    \\

    \begin{center}
    \begin{minipage}[t]{.25\textwidth}
    \small
      \begin{itemize}
        \item Hypothesis: The football coach did not see the play being made.
        \item Label: Contradiction
      \end{itemize}
    \end{minipage}
    \end{center}
    &
    \begin{center}
    \begin{minipage}[t]{.25\textwidth}
    \small
      \begin{itemize}
        \item Hypothesis: The football coach missed seeing the play happen.
        \item Label: Contradiction
      \end{itemize}
    \end{minipage}
    \end{center}
    &
    \begin{center}
    \begin{minipage}[t]{.25\textwidth}
    \small
      \begin{itemize}
        \item Hypothesis: A player injured and getting treated by the team doctor.
        \item Label: Neutral
      \end{itemize}
    \end{minipage}
    \end{center}
    &
    \begin{center}
    \begin{minipage}[t]{.25\textwidth}
    \small
      \begin{itemize}
        \item Hypothesis: An athlete is being attended to by the team doctor for an injury.
        \item Label: Neutral
      \end{itemize}
    \end{minipage}
    \end{center}
    \\ \Xhline{2\arrayrulewidth}

    \begin{center}
    \begin{minipage}{.1\textwidth}
    \includegraphics[width=\linewidth, height=\linewidth]{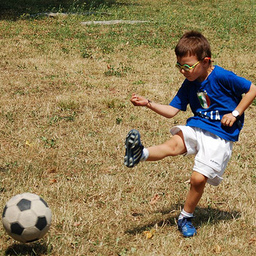}
    \end{minipage}
    \end{center}
    &
    \begin{center}
    \begin{minipage}{.1\textwidth}
    \includegraphics[width=\linewidth, height=\linewidth]{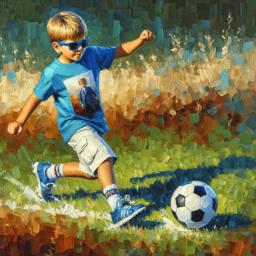}
    \end{minipage}
    \end{center}
    &
    \begin{center}
    \begin{minipage}{.1\textwidth}
    \includegraphics[width=\linewidth, height=\linewidth]{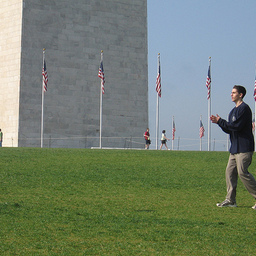}
    \end{minipage}
    \end{center}
    &
    \begin{center}
    \begin{minipage}{.1\textwidth}
    \includegraphics[width=\linewidth, height=\linewidth]{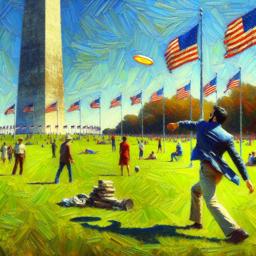}
    \end{minipage}
    \end{center}
    \\

    \begin{center}
    \begin{minipage}[t]{.25\textwidth}
    \small
      \begin{itemize}
        \item Hypothesis: A little boy wearing blue takes a kick at a soccer ball.
        \item Label: Entailment
      \end{itemize}
    \end{minipage}
    \end{center}
    &
    \begin{center}
    \begin{minipage}[t]{.25\textwidth}
    \small
      \begin{itemize}
        \item Hypothesis: A young boy dressed in blue attempts to kick a soccer ball.
        \item Label: Entailment
      \end{itemize}
    \end{minipage}
    \end{center}
    &
    \begin{center}
    \begin{minipage}[t]{.25\textwidth}
    \small
      \begin{itemize}
        \item Hypothesis: The man is in Pakistan kicking a soccer ball.
        \item Label: Contradiction
      \end{itemize}
    \end{minipage}
    \end{center}
    &
    \begin{center}
    \begin{minipage}[t]{.25\textwidth}
    \small
      \begin{itemize}
        \item Hypothesis: The person is playing soccer in Pakistan.
        \item Label: Contradiction
      \end{itemize}
    \end{minipage}
    \end{center}
    \\ \Xhline{3\arrayrulewidth}
    
  \end{tabular}
  }
\end{table}

\newpage
\section{Prompt for Data Annotation}
\label{app:prompt-data-annotation}

In this section, we present an example prompt of the data annotation procedure for the three tasks.

\subsection{Prompt for Data Annotation for Image Captioning Task}

\setlength{\parindent}{0cm}{
\begin{minipage}[t]
{\linewidth}\raggedright

\xhrulefill[thickness=1pt]
\vspace{1mm}
\newline

\texttt{System} \newline
You are an annotator for image captioning tasks. \newline
You will help create stylized image and its captions based on user requests. \newline

\texttt{User} // \textit{Image Decomposition} \newline
Please generate a detailed prompt for DALL-E3 model to replicate the given image. [Original Image] \newline
\texttt{Assistant} \newline
Create an image of a man preparing food outside an industrial-style workspace. The man is wearing a flat cap and a dark short-sleeve shirt and is standing at a brown counter, chopping green onions on a cutting board. Surrounding him on the counter are various fresh vegetables, including green onions, leafy greens, a whole avocado, and a bowl of eggs. In the background, an open garage door reveals the interior of the workspace with tools, a workbench, and a bicycle leaning against the outside. The floor is concrete and the walls are decorated with hanging tools and shelves. The overall atmosphere should convey a casual, industrious vibe. \newline

\texttt{User} // \textit{Style Injection} \newline
Please modify the generated prompt to change the style of the image to a cartoon drawing style. \newline
\texttt{Assistant} \newline
Create a cartoon-style image of a man preparing food outside an industrial-style workspace. The man is wearing a flat cap and a dark short-sleeve shirt and is standing at a brown counter, chopping green onions on a cutting board. Surrounding him on the counter are various fresh vegetables, including green onions, leafy greens, a whole avocado, and a bowl of eggs. In the background, an open garage door reveals the interior of the workspace with tools, a workbench, and a bicycle leaning against the outside. The floor is concrete and the walls are decorated with hanging tools and shelves. The overall atmosphere should convey a casual, industrious vibe, with cartoonish exaggerated features and vibrant colors. \newline

\texttt{User} // \textit{Image Verification} \newline
Please verify if the image below is a cartoon drawing style image of the original image. The generated image should not exactly match the original image but should capture the essence of the original image. Start the response with `Yes' or `No'. [Generated Image] \newline
\texttt{Assistant} \newline
Yes, the provided image is a cartoon-style representation of the original image. The soccer player is depicted in a stylized, animated manner, with exaggerated features typical of cartoons. The attire, including the white short-sleeved jersey, blue shorts, long white socks, and white and orange cleats, closely matches the original image. The animated background with grass, orange cones, and a goal net also retains elements from the original setting, demonstrating a colorful and whimsical portrayal. \newline
\end{minipage}
}

\newpage

\setlength{\parindent}{0cm}{
\begin{minipage}[t]
{\linewidth}\raggedright
\texttt{User} // \textit{Caption Paraphrasing} \newline
Please generate five captions of the generated cartoon drawing style image. The captions should describe the image in different ways. Consider given captions below for reference. \newline
1. a man cutting up vegetables on top of a food cart. \newline
2. A man cutting up scallions at an outdoor table \newline
3. A man chopping vegetables on a white board. \newline
4. A man stands as he cuts up some vegetables. \newline
5. A man stands with a knife and onions in front of a garage \newline
\texttt{Assistant} \newline
1. A cartoon-styled man slicing vegetables at an outdoor table near a garage. \newline
2. An animated character chopping fresh scallions on a cutting board outside. \newline
3. A cartoon figure preparing a meal by cutting vegetables on a white board in front of a workshop. \newline
4. A man in cartoon form stands at an outdoor table, diligently cutting vegetables. \newline
5. A cartoon man, standing with a knife and spring onions, prepares food outside a garage filled with tools.\newline

\vspace{1mm}
\xhrulefill[thickness=1pt]\newline
\end{minipage}
}

\newpage
\subsection{Prompt for Data Annotation for Visual Question Answering Task}
We omitted the image generation process as it is shared across three tasks.

\setlength{\parindent}{0cm}{
\begin{minipage}[t]
{\linewidth}\raggedright

\xhrulefill[thickness=1pt]
\vspace{1mm}
\newline

\texttt{System} \newline
You are an annotator for visual question answering tasks. \newline
You will help create stylized image and its questions based on user requests. \newline

// \textit{Image Generation Process} \newline

\texttt{User} // \textit{Answer Verification} \newline
Please verify if the question and answer pair below is correct for the generated cartoon drawing style image. Start the response with `Yes' or `No'. \newline
Question: Is the person wearing a hat? \newline
\texttt{Assistant} \newline
No, the question and answer pair is not correct. The person in the generated image is not wearing a hat. \newline

\texttt{User} // \textit{Answer Re-annotation} \newline
Please answer the question below based on the generated cartoon drawing style image. Start the response with `Yes' or `No'. \newline
Question: Is the person wearing a hat? \newline
\texttt{Assistant} \newline
No, the person in the generated image is not wearing a hat. \newline

\texttt{User} // \textit{Question Paraphrasing} \newline
Please paraphrase the question below for the generated cartoon drawing style image. The paraphrased question should have the same meaning as the original question but be rephrased in a different way. Only the question should be paraphrased. \newline
Question: Is the person chopping green onions? \newline
\texttt{Assistant} \newline
Paraphrased Question: Is the individual slicing green onions?

\vspace{1mm}
\xhrulefill[thickness=1pt]\newline

\end{minipage}
}

\newpage
\subsection{Prompt for Data Annotation for Visual Entailment Task}


\setlength{\parindent}{0cm}{
\begin{minipage}[t]
{\linewidth}\raggedright

\xhrulefill[thickness=1pt]
\vspace{1mm}
\newline

\texttt{System} \newline
You are an annotator for visual entailment tasks. \newline
You will help create stylized image and its corresponding hypothesis based on user requests. \newline

// \textit{Image Generation Process} \newline

\texttt{User} // \textit{Label Verification} \newline
Please verify if given hypothesis pair and its label is correct for the generated cartoon drawing style image. Start the response with `Yes' or `No'. \newline
Hypothesis: The person is preparing ingredients for a meal in an outdoor kitchen setup. \newline
Label: True \newline
\texttt{Assistant} \newline
No, the hypothesis ``The person is preparing ingredients for a meal in an outdoor kitchen setup'' is not entailed by the given image. \newline

\texttt{User} // \textit{Label Re-annotation} \newline
Does the given hypothesis entail the image? Start the response with 'True', 'False', or 'Undetermined'. \newline
Hypothesis: The person is preparing ingredients for a meal in an outdoor kitchen setup. \newline
\texttt{Assistant} \newline
Undetermined. It is unclear in the generated image if the person is preparing ingredients ``in an outdoor kitchen setup.'' \newline

\texttt{User} // \textit{Hypothesis Paraphrasing} \newline
Please paraphrase the hypothesis sentence below for the generated cartoon drawing style image. The paraphrased hypothesis should have the same meaning as the original sentence but be rephrased in a different way. Only the sentence should be paraphrased.\newline
Hypothesis: The person is preparing ingredients for a meal in an outdoor kitchen setup.\newline
\texttt{Assistant} \newline
Paraphrased Hypothesis: The individual is getting ingredients ready for cooking in an outdoor kitchen setting.

\vspace{1mm}
\xhrulefill[thickness=1pt]\newline

\end{minipage}
}

\newpage

\end{document}